\documentclass[runningheads]{llncs}
\usepackage{graphicx}

\usepackage{tikz}
\usepackage{comment}
\usepackage{amsmath,amssymb} 
\usepackage{color}

\usepackage[width=122mm,left=12mm,paperwidth=146mm,height=193mm,top=12mm,paperheight=217mm]{geometry}

\usepackage{graphicx}
\usepackage{amsmath}
\usepackage{amssymb}
\usepackage{booktabs}

\usepackage{multirow}                 

\usepackage[misc]{ifsym}      

\usepackage{bbding}
\usepackage{pifont}
\usepackage{wasysym}
\usepackage{amssymb}

\usepackage[normalem]{ulem}           
\useunder{\uline}{\ul}{}

\usepackage[capitalize]{cleveref}
\crefname{section}{Sec.}{Secs.}
\Crefname{section}{Section}{Sections}
\Crefname{table}{Table}{Tables}
\crefname{table}{Tab.}{Tabs.}

\begin{document}
\pagestyle{headings}
\mainmatter
\def\ECCVSubNumber{4336}  

\title{Learning Mutual Modulation for \\ Self-Supervised Cross-Modal Super-Resolution} 


\titlerunning{Learning Mutual Modulation for Self-Supervised Cross-Modal SR}
%
\author{Xiaoyu Dong\inst{1,2} \and   
Naoto Yokoya\inst{1,2}$^{(\textrm{\Letter})}$ \and
Longguang Wang\inst{3} \and
Tatsumi Uezato\inst{4}}
\authorrunning{X. Dong et al.}
%
\institute{The University of Tokyo, Tokyo, Japan \and 
RIKEN AIP, Tokyo, Japan \and
National University of Defense Technology, Changsha, China \and
Hitachi, Ltd, Tokyo, Japan \\
\email{dong@ms.k.u-tokyo.ac.jp, yokoya@k.u-tokyo.ac.jp} \\
\url{https://github.com/palmdong/MMSR}}
\maketitle

\begin{abstract} 
Self-supervised cross-modal super-resolution (SR) can overcome the difficulty of acquiring paired training data, but is challenging because only low-resolution (LR) source and high-resolution (HR) guide images from different modalities are available.
Existing methods utilize pseudo or weak supervision in LR space and thus deliver results that are blurry or not faithful to the source modality.  
To address this issue, we present a mutual modulation SR (MMSR) model, which tackles the task by a mutual modulation strategy, including a source-to-guide modulation and a guide-to-source modulation.
In these modulations, we develop cross-domain adaptive filters to fully exploit cross-modal spatial dependency  
and help induce the source to emulate the resolution of the guide and induce the guide to mimic the modality characteristics of the source.   
Moreover, we adopt a cycle consistency constraint to train MMSR in a fully self-supervised manner. 
Experiments on various tasks demonstrate the state-of-the-art performance of our MMSR. 
\end{abstract}


\setcounter{footnote}{0}        

\section{Introduction}
\label{sec:intro}

\begin{figure}[t]
  \centering
   \includegraphics[width=0.88\linewidth]{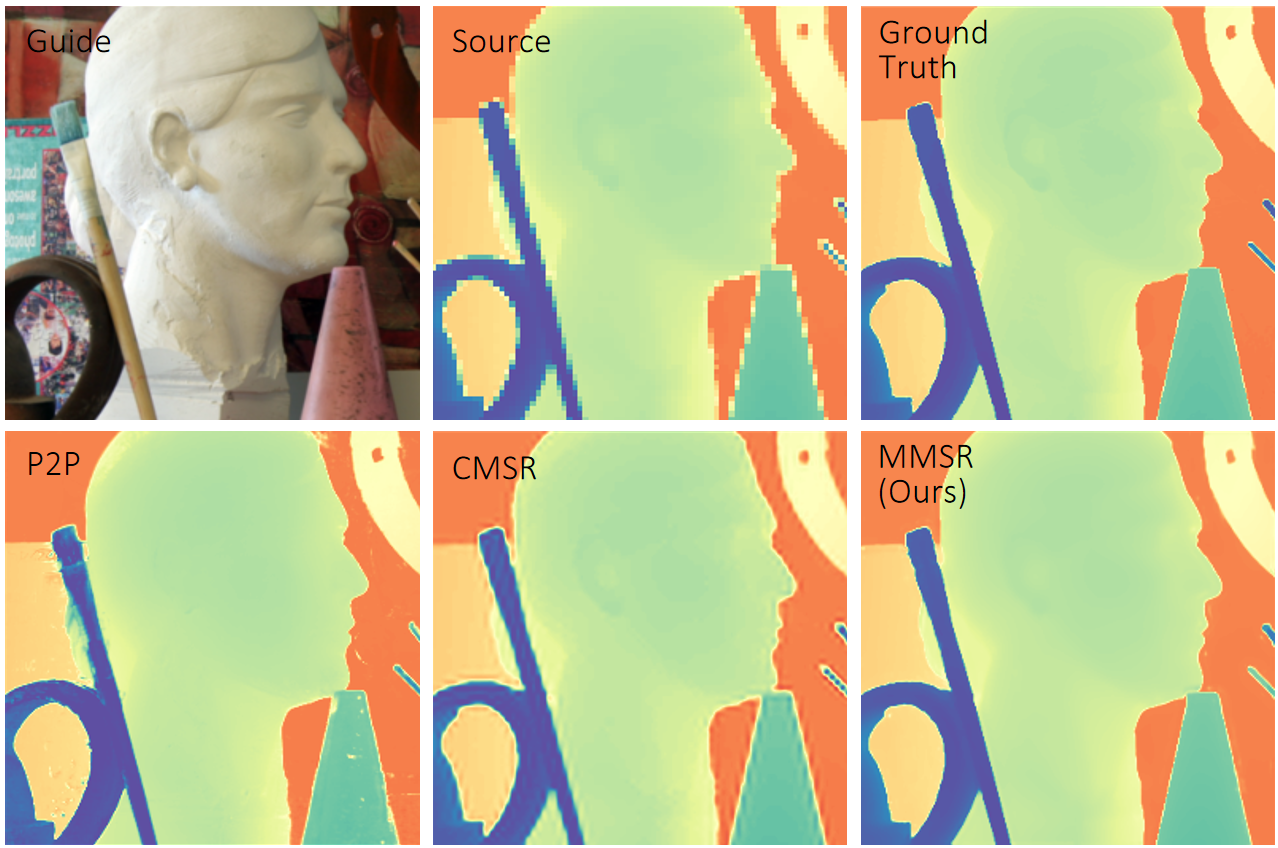}
   \caption{$\times 4$ depth SR results from CMSR~\cite{Shacht_2021_CVPR}, P2P~\cite{Lutio_2019_ICCV}, and our MMSR. Our MMSR achieves results that are HR and faithful to the source modality}
   \label{fig1}
\end{figure}

Multi-modal data, e.g., visible RGB, depth, and thermal, can reflect diverse physical properties of scenes and objects and are widely applied in practice~\cite{Bruno_2021,Silberman_ECCV_2012,Yingqian_2022_CVPR,Arar_2020_CVPR}. 
While high-resolution (HR) visible data is easy to acquire, non-visible modalities are usually low-resolution (LR) due to sensor limitations~\cite{Gu_2020_PAMI,Almasri_2018_ACCV}.
This hinders practical applications and introduces the necessity of cross-modal super-resolution~(SR). 

Cross-modal SR aims at increasing the resolution of an LR modality (source) by using as guidance the structural cues from an HR modality (guide). 
This is difficult due to the spatial discrepancy between different modalities~\cite{Gu_2020_PAMI,Dong_2021_PAMI,Uezato_2020_ECCV}. 

In recent years, deep CNNs have been widely studied to explore source-guide spatial dependency for cross-modal SR, and numerous networks have been developed~\cite{Hui_2016_ECCV,Almasri_2018_ACCV,Li_2019_PAMI,Gu_2020_PAMI,Dong_2021_PAMI,Kim_2021_IJCV,Su_2019_CVPR}. 
However, these methods rely on HR source images~(i.e.,~ground truth) for supervised learning and suffer limited generalization performance, since large-scale training data with paired ground truth is hard to collect~\cite{Lutio_2019_ICCV,He_2021_CVPR,Shacht_2021_CVPR}. 


To address this issue, several efforts~\cite{Shacht_2021_CVPR,Lutio_2019_ICCV} have been made to learn cross-modal SR in a self-supervised\footnote{In this paper, self-supervised learning~\cite{Dong_2022_ECCV} refers to learning from data without paired ground truth in source modality, i.e., only an LR source and an HR guide.} manner.  
Such methods do not require external training data and perform online learning on each combination of LR source and HR guide, thus providing a strong generalization capability. 
However, they face two technical challenges:
\textbf{First},~training SR models using two images that have modality difference and cannot form a direct LR-to-HR mapping.     
\textbf{Second},~achieving high spatial resolution as well as faithful image modality without accessing the supervision from HR source.  
To find a solution, CMSR~\cite{Shacht_2021_CVPR} further downsampled the LR source to generate pseudo paired data in LR space.
P2P~\cite{Lutio_2019_ICCV} formulated the task as a modality transformation problem of the HR guide, 
and employed the LR source as weak supervision. 
While succeeding in training the models, 
these methods cannot overcome the second challenge and deliver results that are blurry or not faithful to the source modality (\cref{fig1}).   
Overall, robust self-supervised cross-modal SR remains an open problem.


In this paper, we tackle self-supervised cross-modal SR by modulating the source and the guide with a cycle consistency constraint (\cref{fig2_model}). 
Specifically, we introduce a mutual modulation strategy,
which includes a source-to-guide modulation (\cref{fig3_s2g}) to induce the source to emulate the resolution of the guide, and a guide-to-source modulation (\cref{fig4_g2s}) to bring the guide closer to the source with respect to the characteristics of the imaging modality.
During the modulations, we develop cross-domain adaptive filters to fully exploit the spatial dependency between the source and the guide and drive our mutual modulation.  
Moreover, we adopt a cycle consistency loss between the downsampled SR result and the original LR source input to train our mutual modulation SR (MMSR) model in a fully self-supervised manner. 
It is demonstrated that our MMSR achieves state-of-the-art performance and produces results with both fine spatial detail and faithful image modality~(\cref{fig1}). 


\textbf{Contributions:}  
\textbf{(1)} We address an open problem in cross-modal SR, and develop a robust self-supervised MMSR model. 
\textbf{(2)} We propose a mutual modulation strategy, and show correlation-based filtering provides an effective inductive bias for deep self-supervised cross-modal SR models. 
\textbf{(3)} We validate our MMSR on depth, digital elevation model~(DEM), and thermal SR, which involve benchmark data, noisy data, and real-world remote sensing data, demonstrating its robustness, generalizability, and applicability.  
\textbf{(4)}~We compare our MMSR with state-of-the-art supervised and self-supervised methods, comprehensively demonstrating both its quantitative and qualitative superiority.


\section{Related Work}
\label{sec:rela}

In this section, we first review several mainstream works in cross-modal SR. Then we discuss techniques that are related to our work, including modulation networks, image filtering, and cycle-consistent learning.

\subsection{Cross-Modal SR} 

Cross-modal SR has evolved from filtering-based~\cite{Yang_2007_CVPR,He_2013_PAMI,Liu_2013_CVPR,Lu_2015_CVPR}, optimization-based~\cite{Diebel_2005_NIPS,Park_2011_ICCV,Ferstl_2013_ICCV}, and dictionary-based methods~\cite{Li_2012_ICME,Kwon_2015_CVPR} to learning-based methods~\cite{Almasri_2018_ACCV,Lutio_2019_ICCV,Gu_2020_PAMI} over the past decades. 
We focus on learning-based methods and review several supervised and self-supervised methods.

Supervised methods, as in other SR tasks~\cite{Wang_2022_PAMI,Yingqian_2022_PAMI,Zhang_2021_PAMI,Zhang_2021_CVPR,Wang_2021g_FPGAN,Tian_2020_CVPR}, have made great progress. 
Early pioneers~\cite{Hui_2016_ECCV,Li_2016_ECCV,Li_2019_PAMI,Almasri_2018_ACCV} have cast the task in a learning-based manner.
Recent works~\cite{Gu_2017_CVPR,Gu_2020_PAMI,Pan_2019_CVPR,Dong_2021_PAMI,Kim_2021_IJCV,Tang_2021_ACMMM} have studied the spatial dependency between the source and guide images.  
Representative work includes the weighted analysis sparse representation model~\cite{Gu_2017_CVPR,Gu_2020_PAMI} and the spatially variant linear representation model~\cite{Pan_2019_CVPR,Dong_2021_PAMI}. 
While obtaining promising performance, these methods suffer limited generalization performance in real-world scenes since large-scale paired training data is hard to acquire~\cite{Lutio_2019_ICCV,He_2021_CVPR,Shacht_2021_CVPR}.  


To address this issue, self-supervised methods without external training have been studied~\cite{Shacht_2021_CVPR,Lutio_2019_ICCV}.
Such methods perform online learning on each combination of LR source and HR guide, and so can be adapted to any given scenario.  
Existing methods conduct the task by forming pseudo supervision in LR space~\cite{Shacht_2021_CVPR} or interpret the task as cross-modal transformation in a weakly supervised manner~\cite{Lutio_2019_ICCV}. While successfully training the models, their delivered results, caused by the non-ideal supervisions, are blurry or not faithful to the source modality.


\subsection{Modulation Networks}  

Modulation networks are emerging in different research fields~\cite{Yang_2018_CVPR,Liu_2021_ICCV,Lee_2021_CVPR_retrieval}.
In image restoration, researchers have developed modulation networks to control restoration capability and flexibility~\cite{He_2019_CVPR,Wang_2019_CVPR,Wang_2019_ICCV}. 
Wang et al.~\cite{Wang_2021_DASR} presents a degradation-aware modulation block to handle different degradations in blind SR. 
Xu et al.~\cite{Xu_2021_CVPR} designs a temporal modulation network to achieve arbitrary frame interpolation in space-time video SR.
In speech separation, Lee et al.~\cite{Lee_2021_CVPR_speech} introduces a cross-modal affinity transformation to overcome the frame discontinuity between audio and visual modalities.
In image retrieval, Lee et al.~\cite{Lee_2021_CVPR_retrieval} introduces content-style modulation to handle the task by exploiting text feedback. 

We propose a mutual modulation strategy to tackle self-supervised cross-modal SR.
Our strategy enables our model to achieve results with both high spatial resolution and faithful image modality, and outperform even state-of-the-art supervised methods.


\subsection{Image Filtering}   
\label{subsec:nonlocal}

Many vision applications involve image filtering to suppress and/or extract content of interests in images~\cite{He_2013_PAMI}.
Simple linear filters have been extensively used in image tasks such as sharpening, stitching~\cite{Perez_2003_ACM}, and matting~\cite{Sun_2004_ACM}. 
In the area of image restoration, local~\cite{Rudin_1994_ICIP} and non-local~\cite{Burger_2012_CVPR,Dabov_2007_TIP} filtering have been studied. 
Liu~et~al.~\cite{Liu_2018_NLRN} first incorporated non-local operations in a neural network for denoising and SR.  
Later researchers used non-local layers to exploit the self-similarity prior for restoration quality improvement~\cite{zhang2019rnan,Mei_2020_CVPR,Mei_2021_CVPR,Shim_2020_CVPR}.

Differently, we aim at handling multi-modal images that have local spatial dependency~\cite{Dong_2021_PAMI,Pan_2019_CVPR}, discrepancy, and resolution gap.  
Therefore, we learn filters confined to pixel neighborhoods across features from the source and guide modality domains to exploit the local spatial dependency of different modalities and drive our mutual modulation. 
Experiments in~\Cref{subsec:ablation} show that our filters can eliminate the spatial discrepancy and resolution gap of multi-modal images, providing an effective inductive bias for cross-modal SR models.


\subsection{Cycle-Consistent Learning}  

Given a data pair $A$ and $B$, cycle-consistent learning aims to train deep models by establishing a closed cycle with a forward $A$-to-$B$ mapping and a backward $B$-to-$A$ mapping. 
This idea has been investigated in vision tasks such as visual tracking~\cite{Kalal_2010_ICPR,Sundaram_2010_ECCV}, dense semantic alignment~\cite{Zhou_2015_CVPR,Zhou_2016_CVPR}, and image translation~\cite{CycleGAN2017,Zhang_2020_CVPR}. 
In image restoration, researchers imposed cycle consistency constraint to image dehazing~\cite{Shao_2020_CVPR} and unpaired SR~\cite{Maeda_2020_CVPR}.

We introduce cycle-consistent learning to cross-modal SR, and adopt a cycle consistency loss to encourage the downsampled SR source and the original LR source input to be consistent with each other. 
This allows our model to be trained in a fully self-supervised manner. 


\begin{figure*}[!t]
  \centering
  \includegraphics[width=1\linewidth]{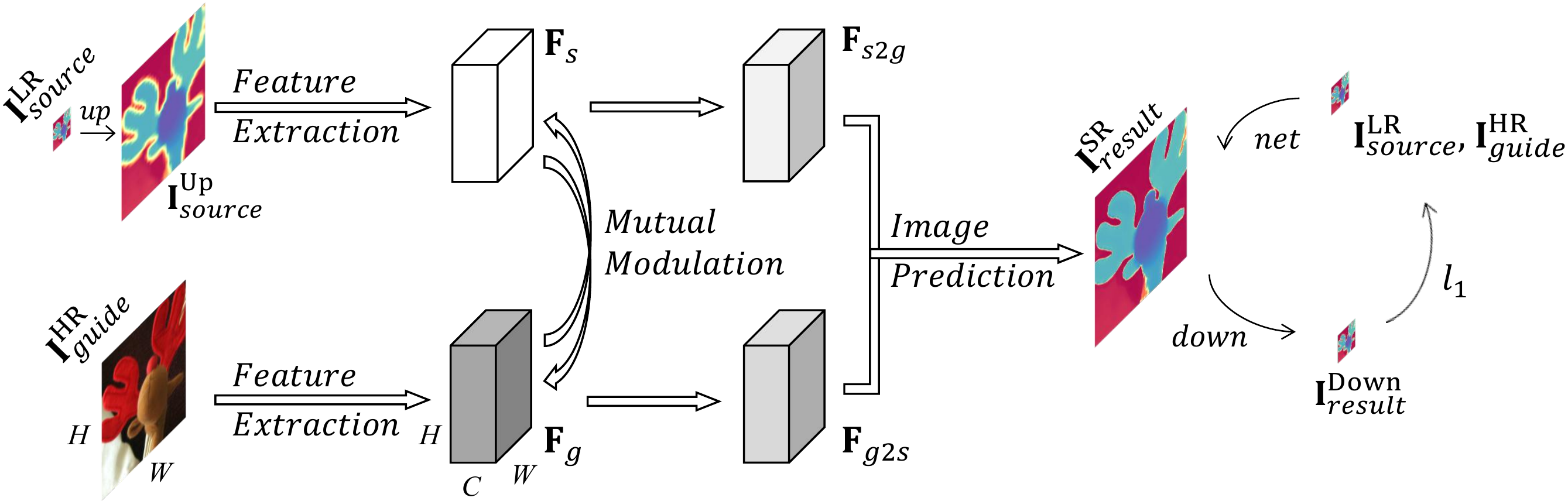} 
  \caption{An illustration of our MMSR model. 
  During mutual modulation, the source is induced to emulate the resolution of the guide, while the guide is induced to mimic the characteristics of the source.  
  A cycle consistency constraint is adopted to conduct training in a fully self-supervised manner. 
  $'up'$, $'net'$, $'down'$, and $'l_1'$ represent upsampling, our network, downsampling, and loss term, respectively}  
  \label{fig2_model}
\end{figure*}

\section{Method}
\label{sec:method}

As illustrated in~\cref{fig2_model}, our MMSR starts from an LR source $\textbf{I}_{source}^{\rm LR}$ and the HR guide $\textbf{I}_{guide}^{\rm HR}$.  
It then modulates the source feature $\textbf{F}_{s}$ extracted from the bilinearly upsampled source $\textbf{I}_{source}^{\rm Up}$,
which still presents low resolution and lacks of spatial detail, and the guide feature $\textbf{F}_{g}$ extracted from $\textbf{I}_{guide}^{\rm HR}$, which contains HR structural cues important to the source and also discrepancy patterns. 
Finally, it predicts the SR source $\textbf{I}_{result}^{\rm SR}$ from the fusion of the modulated $\textbf{F}_{s2g}$ and $\textbf{F}_{g2s}$, and constrains itself by casting $\textbf{I}_{result}^{\rm SR}$ back to $\textbf{I}_{source}^{\rm LR}$. 


\subsection{Mutual Modulation} 
\label{sec_modulation}

In our mutual modulation, $\textbf{F}_s$ and $\textbf{F}_g$ are optimized by taking each other as reference.  
Cross-domain adaptive filters are developed as basic operators to drive the modulation.
 
\begin{figure}[t]
  \centering
   \includegraphics[width=0.53\linewidth]{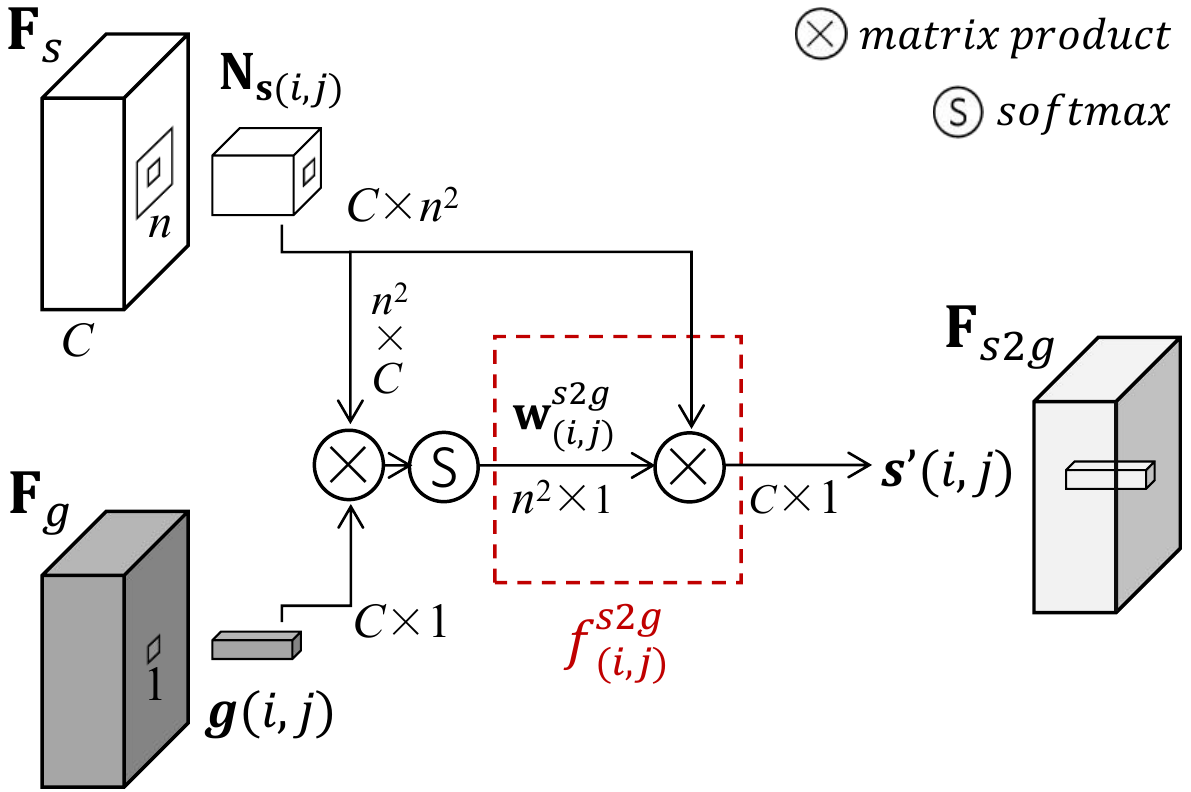}
   \caption{An illustration of the source-to-guide modulation. Filters update the pixels in $\textbf{F}_s$ by targeting the counterparts in $\textbf{F}_g$ to induce $\textbf{F}_{s}$ to become HR} 
   \label{fig3_s2g}
\end{figure}

In the \textit{\textbf{Source-to-Guide Modulation}} (\cref{fig3_s2g}), $\textbf{F}_s$ is modulated to emulate the high resolution of $\textbf{F}_g$. 
Specifically, to each pixel in $\textbf{F}_s$ (denoted as $\textbf{s}_{(i,j)}$),  
we learn a filter $f^{s2g}_{(i,j)}(\cdot)$ confined to its neighbor pixels in an $n\times n$ neighborhood (denoted as $\textbf{N}_{\textbf{s}_{(i,j)}}$)) 
and target its counterpart pixel in $\textbf{F}_g$ (denoted as $\textbf{g}_{(i,j)}$)\footnote{$\textbf{s}_{(i,j)}$ or $\textbf{g}_{(i,j)}$ denotes the pixel at the $i$-th row and $j$-th column in $\textbf{F}_s$ or $\textbf{F}_g$ and is a vector of size $C\times 1$. $\textbf{N}_{\textbf{s}_{(i,j)}}$ contains $n\times n$ pixels and is a tensor of shape $C\times n\times n$.}.
The filter weight is calculated as\footnote{$\textbf{N}_{\textbf{s}_{(i,j)}}$ is first reshaped to a $C\times n^2$ matrix. Matrix multiplication is then taken between the transpose of the matrix and $\textbf{g}_{(i,j)}$, which results in a $n^2\times 1$ vector. Filter weight $\textbf{w}^{s2g}_{(i,j)}$ is obtained by taking a softmax normalization to the resulting vector and is also of size $n^2\times 1$.}:  
\begin{equation}
  \textbf{w}^{s2g}_{(i,j)} = {\rm softmax} \Big( \big(\textbf{N}_{\textbf{s}_{(i,j)}} \big)^{\rm T} \textbf{g}_{(i,j)} \Big),
  \label{eq_ws2g}
\end{equation}
and evaluates the correlation value between $\textbf{g}_{(i,j)}$ and each pixel in $\textbf{N}_{\textbf{s}_{(i,j)}}$~\cite{Liu_2018_NLRN,Buades_2005_CVPR}.
Thus our filters allow fully exploitation of the local dependency between the source and guide modalities.
In~\Cref{subsec:ablation}, we experimentally show that such adaptive filters with learning cross-domain correlations can deliver product that is spatially approaching a given target from a different domain, and are effective modulators to $\textbf{F}_s$ and $\textbf{F}_g$ in a case in which neither an HR source feature nor a guide feature without spatial discrepancy is available.  
Here, a filtering operation is expressed as\footnote{$\textbf{w}^{s2g}_{(i,j)}$ weights the reshaped $\textbf{N}_{\textbf{s}_{(i,j)}}$ by taking matrix multiplication to result in a $C\times 1$ vector, i.e., $\textbf{s}'_{(i,j)}$.}:   
\begin{equation}         
  \textbf{s}'_{(i,j)} = f^{s2g}_{(i,j)} \big(\textbf{N}_{\textbf{s}_{(i,j)}} \big) = \textbf{N}_{\textbf{s}_{(i,j)}} \textbf{w}^{s2g}_{(i,j)}, 
  \label{eq_s'}
\end{equation}  
where the resulting $\textbf{s}'_{(i,j)}$ is the update of $\textbf{s}_{(i,j)}$ and is induced to spatially emulate $\textbf{g}_{(i,j)}$, which is in HR domain.   
The whole source-to-guide modulation is conducted by updating all the pixels in $\textbf{F}_s$ by targeting the counterpart guide pixels, resulting in $\textbf{F}_{s2g}$, which inherits the HR property of the guide.

\begin{figure}[t]
  \centering
   \includegraphics[width=0.53\linewidth]{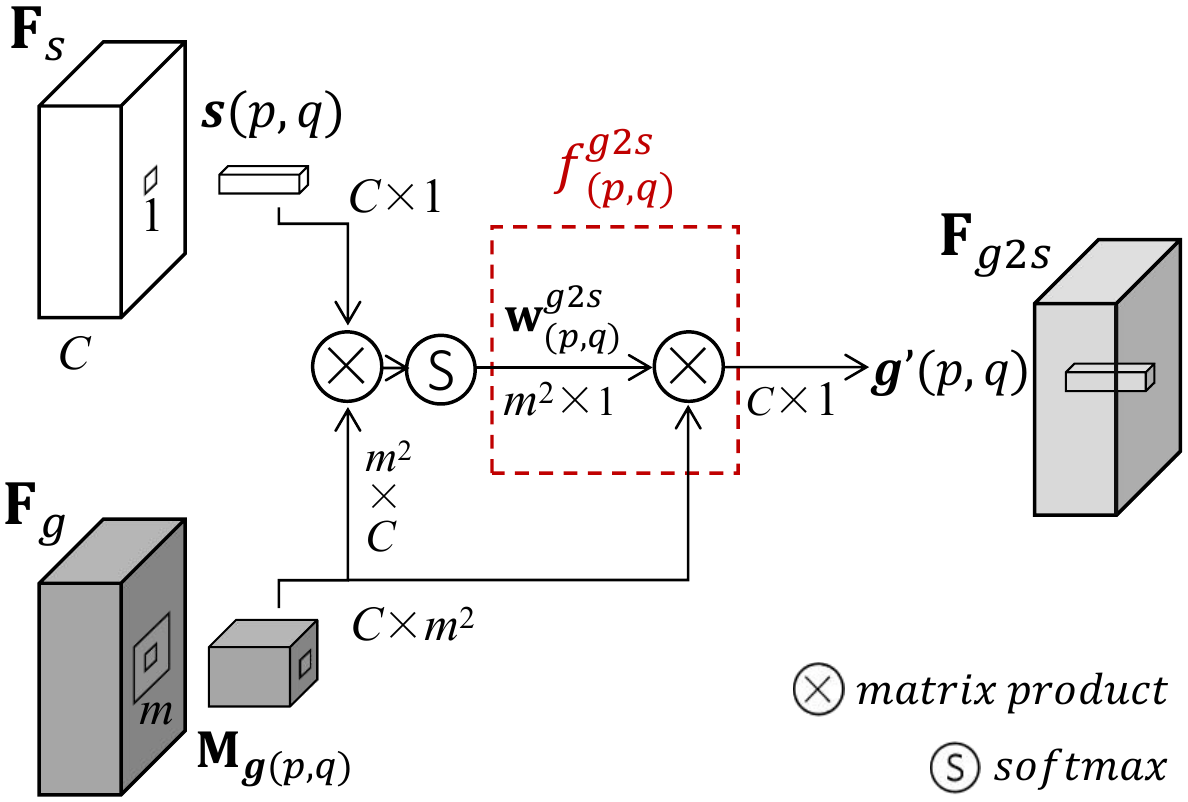}
   \caption{An illustration of the guide-to-source modulation. Filters update the pixels in $\textbf{F}_g$ by targeting $\textbf{F}_s$ to bring $\textbf{F}_g$ closer to the source on modality characteristics} 
   \label{fig4_g2s}
\end{figure}

The \textit{\textbf{Guide-to-Source Modulation}} (\cref{fig4_g2s}) suppresses the discrepancy in $\textbf{F}_g$ to make its characteristics more like those of $\textbf{F}_s$.
To guide pixel $\textbf{g}_{(p,q)}$, we learn a filter $f^{g2s}_{(p,q)}(\cdot)$ specific to the neighbor pixels in an $m\times m$ neighborhood $\textbf{M}_{\textbf{g}_{(p,q)}}$ and
target the source pixel $\textbf{s}_{(p,q)}$. 
The filter weight measuring the cross-domain correlation between $\textbf{M}_{\textbf{g}_{(p,q)}}$ and $\textbf{s}_{(p,q)}$ is calculated as:
\begin{equation}
  \textbf{w}^{g2s}_{(p,q)} = {\rm softmax} \Big( \big(\textbf{M}_{\textbf{g}_{(p,q)}} \big)^{\rm T} \textbf{s}_{(p,q)} \Big). 
  \label{eq_wg2s}
\end{equation}  
$\textbf{g}_{(p,q)}$ is updated as:  
\begin{equation}         
  \textbf{g}'_{(p,q)} = f^{g2s}_{(p,q)} \big(\textbf{M}_{\textbf{g}_{(p,q)}} \big) = \textbf{M}_{\textbf{g}_{(p,q)}} \textbf{w}^{g2s}_{(p,q)}. 
  \label{eq_g'}
\end{equation}  
Updating the guide pixels by considering the correlation to the pixels from the source modality domain allows our model to recognize which patterns in the guide are highly relevant to the source.
Thus the guide-to-source modulation can adaptively suppress the discrepancy patterns in $\textbf{F}_{g}$, resulting in $\textbf{F}_{g2s}$, which has modality characteristics that are close to the source and the structural cues necessary to super-resolve the source.

Ablation studies in~\Cref{subsec:ablation} demonstrate that both the mutual modulation and the cross-domain adaptive filtering play critical roles in developing a model that can yield results with high spatial resolution and faithful image modality.  

\subsection{Cycle-Consistent Self-Supervised Learning} 
\label{sec_loss}

One technical challenge in self-supervised cross-modal SR is using source and guide images that cannot form direct LR-to-HR mapping to train SR models. 

We argue the SR result should stay fully in the source modality domain, 
and therefore train our model with a cycle consistency constraint in which the start is the LR source along with the HR guide, while the end is still the LR source, 
as illustrated in~\cref{fig2_model}.
In the forward mapping, our network works as a regularizer that optimizes both the source and the guide to make a prediction induced to reach the guide in terms of spatial resolution and be faithful to the source in terms of image modality. 
In the backward mapping, we incentivize the consistency between the downsampled prediction and the original LR source input by minimizing $l_1$ norm:
\begin{equation}
  \mathbb{C} = \Big\Vert f_{down} \big(f_{net}(\textbf{I}_{source}^{\rm LR},\textbf{I}_{guide}^{\rm HR}) \big) - \textbf{I}_{source}^{\rm LR} \Big\Vert _1 ,
  \label{eq_loss}
\end{equation}
where $f_{net}$ denotes our network and $f_{down}$ denotes average pooling downsampling.
Our mutual modulation strategy enables our MMSR to successfully avoid a trivial solution of~\Cref{eq_loss}, i.e., an identity function for $f_{net}$. 
Experimental support is provided in~\Cref{subsec:ablation}.

Unlike other self-supervised methods that utilize pseudo or weak supervision in LR space, 
our model starts from the source modality, seeks an optimal prediction in HR space, and then constrains itself by getting back the start. 
In this way, both high resolution and faithful modality can be achieved and the whole process is fully self-supervised.  


\section{Experiments}   
\label{sec:experiments}

\subsection{Experimental Settings}  

\noindent\textbf{Network Architecture.} 
We adopt conventional convolution layers and the residual block from ~\cite{Lim_2017_CVPRW} to construct our network.
The feature extraction branch of the source image (source branch) and the feature extraction branch of the guide image (guide branch) each consists of two convolution layers and two residual blocks.  
The image prediction part contains three residual blocks and one convolution layer. 
In the source branch and the prediction part, the convolution kernel size is set as $1\times1$.  
In the guide branch, the kernel size is $3\times3$.
Before the prediction part, a $1\times1$ convolution is adopt to fuse $\textbf{F}_{s2g}$ and $\textbf{F}_{g2s}$.
The number of channels is 1 for the first convolution in source branch and the last convolution in prediction part; is 3 for the first convolution in guide branch; is 64 for the other convolutions.\medskip\\ 
\noindent\textbf{Implementation Details.} 
We implement our MMSR model with PyTorch on an NVIDIA RTX 3090 GPU, 
and train it through the cycle consistency loss in~\Cref{eq_loss} for 1000 epochs on each combination of LR source and HR guide. 
Adam optimizer~\cite{Kingma_2015_ICLR} is employed.
The learning rate starts from 0.002 and decays by 0.9998 every 5 epochs.  
We do not use data augmentation techniques~\cite{Shacht_2021_CVPR}.\medskip\\ 
\noindent\textbf{Comparison Methods.} 
We compare MMSR with five state-of-the-art cross-modal SR methods, including three supervised methods (FDSR~\cite{He_2021_CVPR}, DKN \cite{Kim_2021_IJCV}, and FDKN~\cite{Kim_2021_IJCV}) and two self-supervised methods (CMSR~\cite{Shacht_2021_CVPR} and P2P~\cite{Lutio_2019_ICCV}).
We implement these methods fully following the settings suggested in their papers.\medskip\\ 
\noindent\textbf{Datasets and Evaluation Metric.} 
We conduct experiments on depth, DEM, and thermal modalities.
For depth SR, we sample three test sets from the widely used Middlebury 2003~\cite{Scharstein_2003_CVPR}, 2005~\cite{Scharstein_2007_CVPR}, and 2014~\cite{Scharstein_2014_GCPR} benchmarks.
These three test sets include 14, 37, and 43 visible-depth pairs of size $320 \times320$, respectively.
The three supervised comparison methods are trained on 1000 data pairs from the NYU v2 benchmark~\cite{Kohli_2012_ECCV}.
For DEM SR, we choose the remote sensing data used in the 2019 IEEE GRSS Data Fusion Contest (DFC)~\cite{DFC_2019_DEM} and create a test set that includes 54 visible-DEM pairs. 
We train the supervised methods on 1000 data pairs.
We follow the protocols in~\cite{de_Lutio_2022_CVPR,Lutio_2019_ICCV} and adopt pooling to generate LR depth and DEM.  
For thermal SR, we use the visible and thermal hyperspectral remote sensing data from the 2014 IEEE GRSS DFC~\cite{DFC_2014_thermal},
and select one band from the original thermal hyperspectral imagery as LR source.
As the evaluation metric, we use the Root Mean Squared Error (RMSE).  


\subsection{Evaluation on Benchmark Depth Data}
\label{subsec:comparison}

\setlength{\tabcolsep}{4pt}
\begin{table}[t]
\begin{center}
\caption{Depth SR on the Middlebury 2003, 2005, and 2014 datasets. We report the average RMSE. The best and the second best results are in \textcolor{red}{red} and \textcolor{blue}{blue},~respectively} 
\label{table1_depth}
\begin{tabular}{cc|ccc|ccc}
\toprule
                       &                & \multicolumn{3}{c|}{Supervised}            &\multicolumn{3}{c}{Self-Supervised}                            \\ \hline
Dataset                & Scale          & DKN~\cite{Kim_2021_IJCV}  & FDKN~\cite{Kim_2021_IJCV} & FDSR~\cite{He_2021_CVPR}   & P2P~\cite{Lutio_2019_ICCV}     & CMSR~\cite{Shacht_2021_CVPR}                   & Ours                \\ \midrule 
                       & $\times 4$     & 2.11 & 1.84 & {\color[HTML]{3531FF} 1.83} & 2.94                        & 2.52 & {\color[HTML]{FE0000} 1.78} \\ \cline{2-8} 
\multirow{-2}{*}{2003} & $\times 8$     & 2.71 & 2.74 & {\color[HTML]{FE0000} 2.55} & 3.03                        & - & {\color[HTML]{3531FF} 2.63} \\ \midrule 
                       & $\times 4$     & 3.14 & 2.79 & {\color[HTML]{3531FF} 2.74} & 3.78                        & 3.51 & {\color[HTML]{FE0000} 2.47} \\ \cline{2-8} 
\multirow{-2}{*}{2005} & $\times 8$     & 4.45 & 4.52 & {\color[HTML]{000000} 4.27} & {\color[HTML]{3531FF} 3.99} & - & {\color[HTML]{FE0000} 3.92} \\ \midrule 
                       & $\times 4$     & 2.88 & 2.51 & {\color[HTML]{3531FF} 2.41} & 3.90                        & 2.87 & {\color[HTML]{FE0000} 2.30} \\ \cline{2-8} 
\multirow{-2}{*}{2014} & $\times 8$     & 4.21 & 4.06 & {\color[HTML]{3531FF} 4.00} & 4.13                        & - & {\color[HTML]{FE0000} 3.60}\\\bottomrule 
\end{tabular}
\end{center}
\end{table}
\setlength{\tabcolsep}{1.4pt}
\begin{figure*}[t]
  \centering
  \includegraphics[width=1\linewidth]{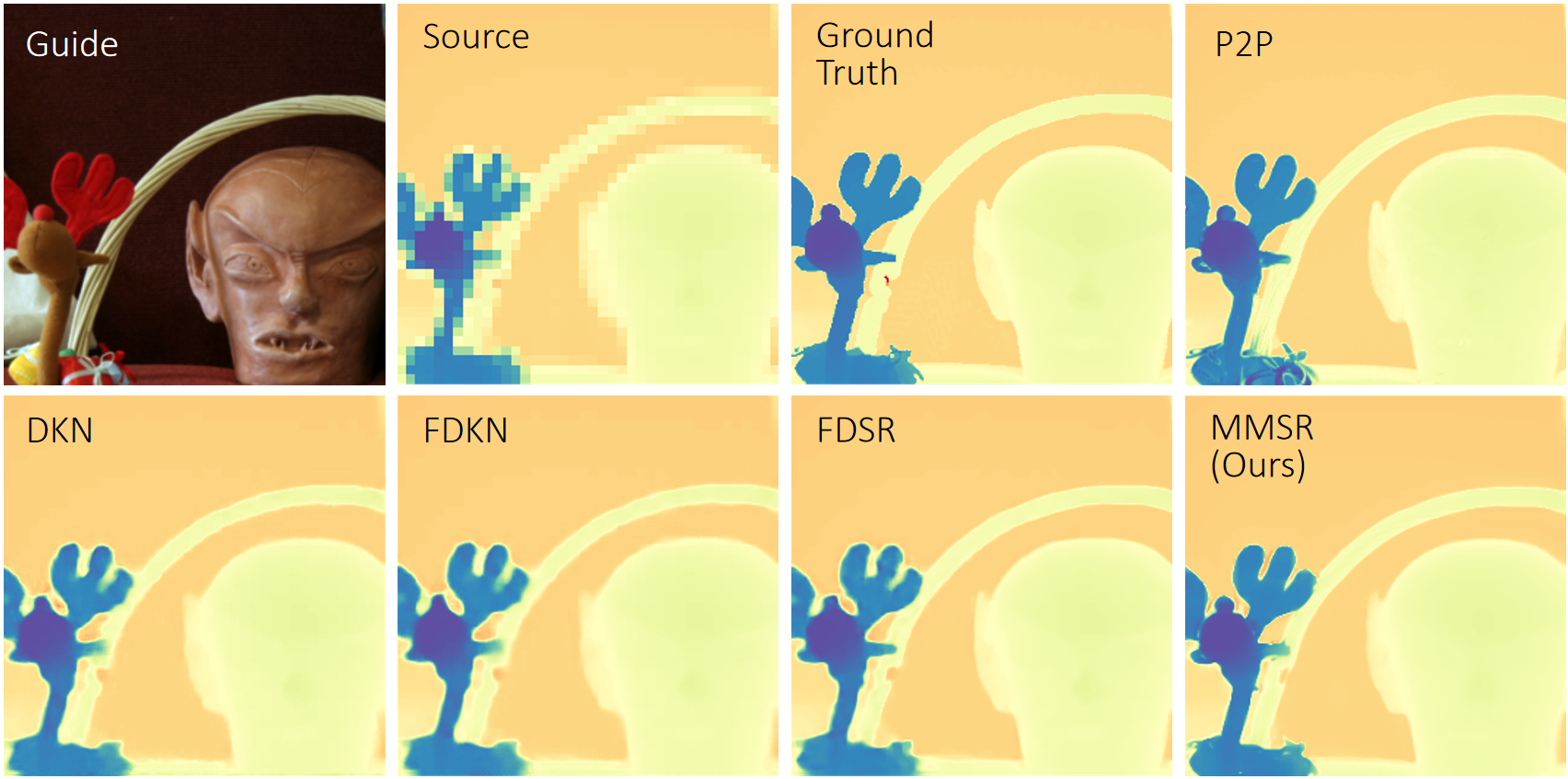} 
  \caption{$\times 8$ depth SR results on the Middlebury 2005 dataset. All the compared methods~\cite{Shacht_2021_CVPR,He_2021_CVPR,Kim_2021_IJCV}, except for P2P~\cite{Lutio_2019_ICCV}, take both LR source and HR guide as input} 
  \label{fig5_2005}
\end{figure*} 

\begin{figure*}
  \centering
  \includegraphics[width=1\linewidth]{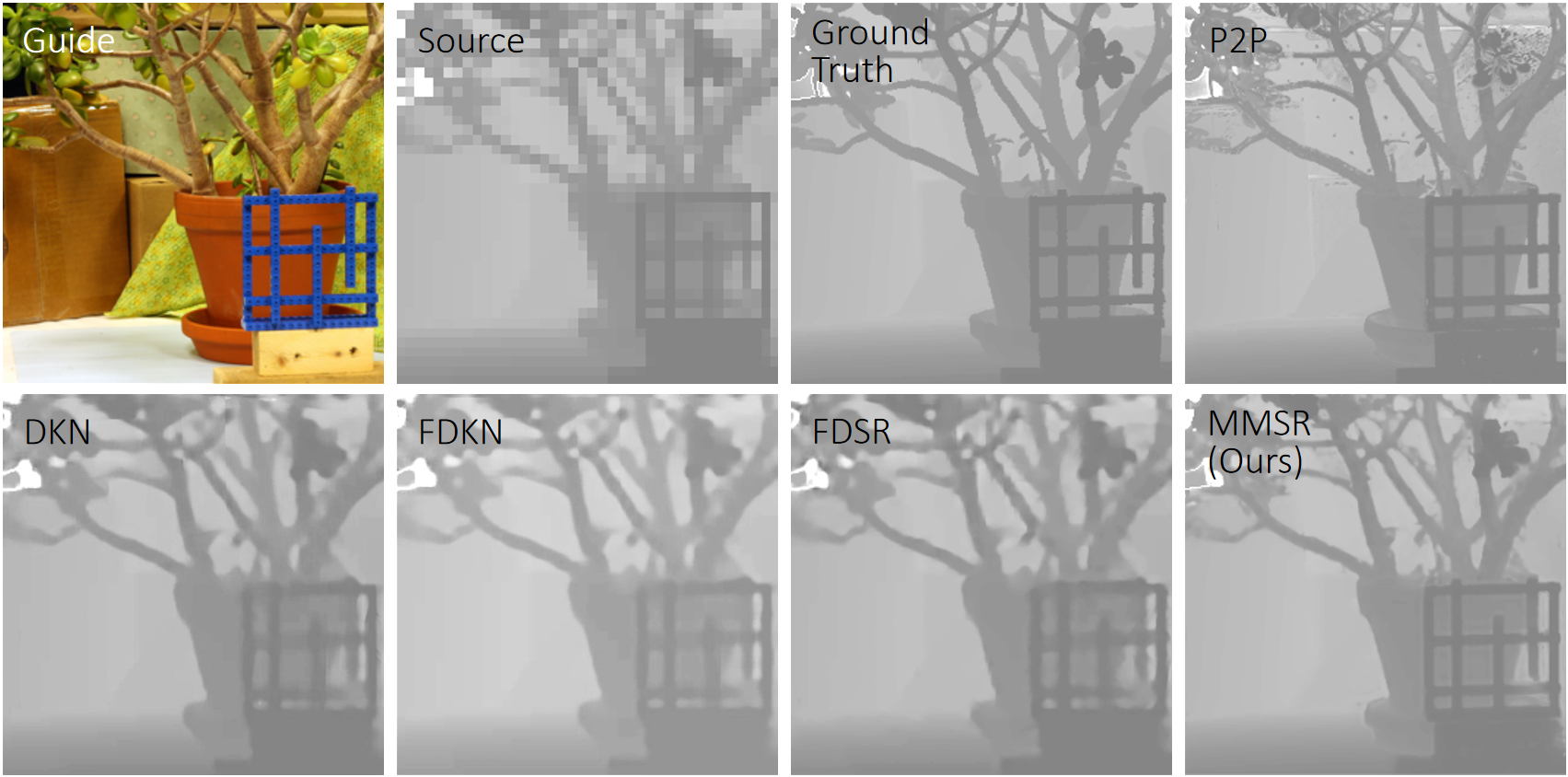}  
  \caption{$\times 8$ depth SR results on the Middlebury 2014 dataset}
  \label{fig6_2014}
\end{figure*}  

\begin{figure}[t]
  \centering
   \includegraphics[width=0.96\linewidth]{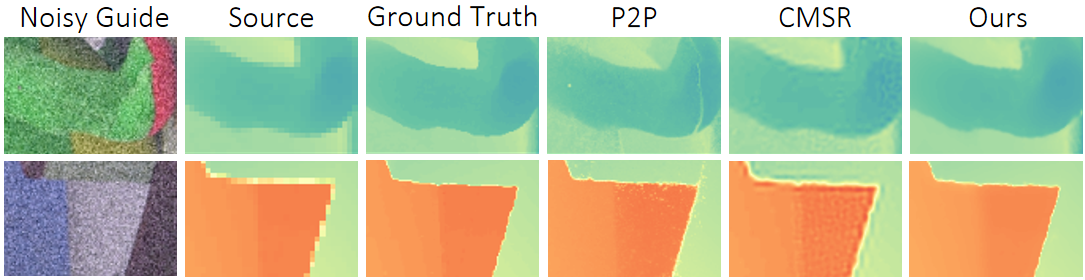} 
   \caption{$\times 4$ depth SR on the Middlebury 2003 dataset under noisy guidance. $'$Noisy Guide$'$ is generated by adding Gaussian noise with noise level 50 
   }
   \label{fig7_noisy}
\end{figure}

In this section, we compare MMSR with five state-of-the-art methods~\cite{Shacht_2021_CVPR,He_2021_CVPR,Kim_2021_IJCV,Lutio_2019_ICCV} on the Middlebury benchmarks.
In the source-to-guide modulation of MMSR, the neighborhood size for filtering is set as $11 \times11$.
In the guide-to-source modulation, it is $5 \times5$. 
The effect of the neighborhood size is analyzed in~\Cref{subsec:ablation}.

\Cref{table1_depth} quantitatively reports $\times 4$ and $\times 8$ SR results. 
We do not provide the $\times 8$ SR results of CMSR~\cite{Shacht_2021_CVPR}
because its training settings for high scale factors is not reported in its paper.  
As we can see, our MMSR consistently outperforms previous self-supervised methods~\cite{Shacht_2021_CVPR,Lutio_2019_ICCV}, 
as well as fully supervised methods~\cite{He_2021_CVPR,Kim_2021_IJCV} that are trained under the supervision from HR source.

\cref{fig5_2005} and \cref{fig6_2014} visualize $\times 8$ SR results on the Middlebury 2005 and 2014 datasets. 
We can observe that performing cross-modal SR as weakly-supervised cross-modal transformation allows P2P~\cite{Lutio_2019_ICCV} to maintain the resolution of the guide yet incurs serious discrepancy artifacts. 
FDSR~\cite{He_2021_CVPR}, DKN~\cite{Kim_2021_IJCV}, and FDKN~\cite{Kim_2021_IJCV} produce results that are faithful to the source modality but spatially blurry,
because supervised methods cannot easily generalize well the test data.  
In contrast, our MMSR does not require external training and optimizes both the source and the guide with a cycle-consistent constraint, thus achieving strong generalization performance and resulting in both high spatial resolution and faithful modality characteristics.  

In \cref{fig7_noisy}, we further compare our MMSR with the two self-supervised methods (CMSR~\cite{Shacht_2021_CVPR} and P2P~\cite{Lutio_2019_ICCV}) to study their robustness to noisy guidance.   
Thanks to our mutual modulation strategy which filters and updates multi-modal inputs by considering their correlation at a pixel level, our MMSR shows stronger robustness to guide images with heavy noise.  


\setlength{\tabcolsep}{4pt}
\begin{table}[!h]
\begin{center}
\caption{Effectiveness study of mutual modulation strategy}
\label{table2_modulation}
\begin{tabular}{c|c|c|c|c}
\toprule
                & Model$_0$   & Model$_1$    & Model$_2$    & Model$_3$ \\ \midrule
Source-to-Guide & \ding{55}   & \ding{55}  & \checkmark    & \checkmark    \\ \hline
Guide-to-Source & \ding{55}   & \checkmark & \ding{55}    & \checkmark    \\ \midrule
RMSE            & 4.30  & 4.08 & 3.72 & 3.67 \\ \bottomrule 
\end{tabular}
\end{center}
\end{table}
\setlength{\tabcolsep}{1.4pt}
\begin{figure}
  \centering
   \includegraphics[width=0.9\linewidth]{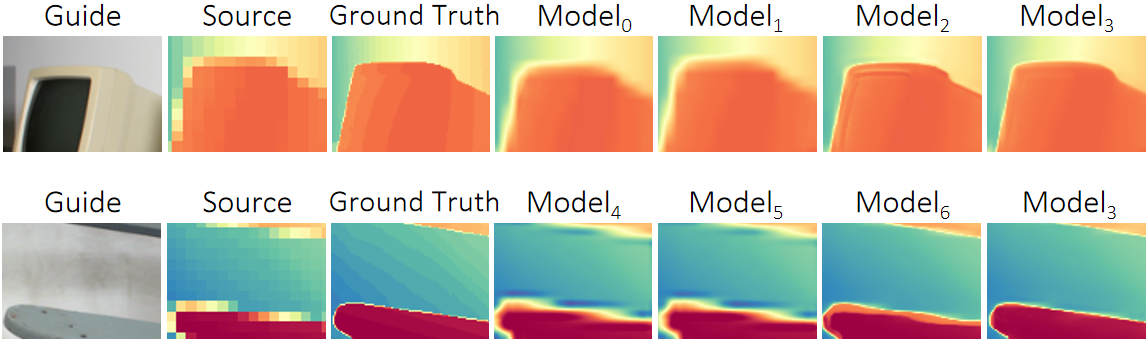} 
   \caption{SR results from models with different modulation settings (upper row) or variant filters (lower row)}
   \label{fig8_modulation}
\end{figure}

\subsection{Ablation Study} 
\label{subsec:ablation}

We analyze MMSR by observing $\times 8$ SR results on the Middlebury 2014 dataset. 

\noindent\textbf{Mutual Modulation Strategy.} 
We first clarify the effectiveness and necessity of the proposed source-to-guide and guide-to-source modulations with setting the both neighborhood sizes for filtering as $11\times 11$.
\Cref{table2_modulation} reports models that adopt different modulation settings.  
Visual examples are shown in \cref{fig8_modulation}~(upper row). 
We can see Model$_0$ without mutual modulation tends toward the unimproved solution of~\Cref{eq_loss}.
Model$_1$ with only the guide-to-source modulation also produces blurred results since the input source stays unimproved.
Model$_2$ yields better performance as the source-to-guide modulation increases the resolution of the source. 
By performing mutual modulation between source and guide, Model$_3$ successfully overcomes the challenge that limits existing self-supervised methods, and yields results that are HR and faithful to the source modality, though no any supervision from ground truth is given. 

\begin{figure}[!t]
  \centering
   \includegraphics[width=0.88\linewidth]{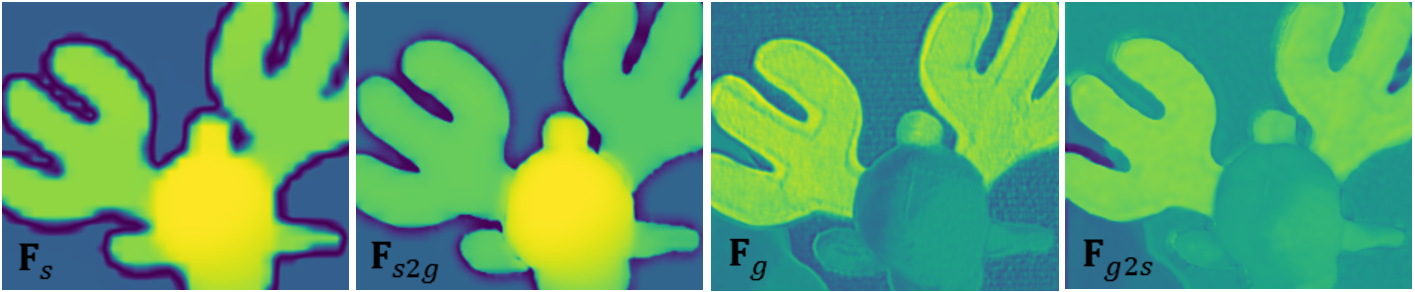} 
   \caption{Visualization of features before and after cross-domain adaptive filtering 
   } 
\label{fig9_feature}
\end{figure}

\noindent\textbf{Cross-Domain Adaptive Filtering.} 
Based on Model$_3$, we highlight our developed filters with regard to two factors: 
\textbf{(1)}~cross-domain learning (our filters measure the correlation across the source and guide modality domains) and 
\textbf{(2)}~adaptiveness (our filter weights are variant for different pixel neighborhoods).
\Cref{table3_filter}~compares Model$_3$ and three models with variant filters.  
Visual examples are in \cref{fig8_modulation}~(lower row).
Note that, in Model$_3$, our filters $f^{s2g}_{(i,j)}$ and $f^{g2s}_{(p,q)}$ are of size $11\times 11$.
In Model$_4$, we replace $f^{s2g}_{(i,j)}$ with an $11\times 11$ convolution to filter $\textbf{F}_{s}$, 
replace $f^{g2s}_{(p,q)}$ with another $11\times 11$ convolution to filter $\textbf{F}_{g}$, 
and fuse the filtering products using a $1\times 1$ convolution. 
In Model$_5$, we first concatenate $\textbf{F}_{s}$ and $\textbf{F}_{g}$, then fuse them using an $11\times 11$ convolution.  
Due to insufficient consideration of the dependency between source and guide domains
and the weight invariance of conventional convolutions~\cite{Esser_2021_CVPR}, 
Model$_4$ and Model$_5$ show inferior performance. 
In Model$_6$, we change the target pixel of $f^{s2g}_{(i,j)}(\cdot)$ (i.e., $\textbf{g}_{(i,j)}$) to $\textbf{s}_{(i,j)}$, and change the target pixel of $f^{g2s}_{(p,q)}(\cdot)$ (i.e., $\textbf{s}_{(p,q)}$) to $\textbf{g}_{(p,q)}$, resulting in adaptive filters $f^{s2s}_{(i,j)}(\cdot)$ and $f^{g2g}_{(p,q)}(\cdot)$ similar to non-local filtering~\cite{Liu_2018_NLRN}.
As we can observe, Model$_6$ gets improvement by suppressing artifacts caused by bilinear interpolation,
but is still inferior as its filters cannot update $\textbf{F}_{s}$ and $\textbf{F}_{g}$ properly due to the no measurement of cross-domain correlations.
In Model$_3$, 
our cross-domain adaptive filters $f^{s2g}_{(i,j)}(\cdot)$ and $f^{g2s}_{(p,q)}(\cdot)$ fully exploit cross-modal spatial dependency,
update $\textbf{F}_{s}$ by considering pixel correlation to $\textbf{F}_{g}$ from the HR guide domain, 
and update $\textbf{F}_{g}$ by considering pixel correlation to $\textbf{F}_{s}$ from the source modality domain.  
\cref{fig9_feature}~visualizes features before and after filtering.  
Our filters optimize the resolution of the source feature and suppress the discrepancy of the guide feature,
enabling an effective inductive bias and the superior performance of Model$_3$.

\setlength{\tabcolsep}{4pt}
\begin{table}[!t] 
\begin{center}
\caption{Effectiveness study of cross-domain adaptive filters} 
\label{table3_filter}
\begin{tabular}{c|c|c|c|c}
\toprule
           & Model$_4$ & Model$_5$ & Model$_6$ & Model$_3$ \\ \midrule
Cross-Domain & \ding{55} & \textbf{\checkmark}    & \ding{55} & \textbf{\checkmark}      \\ \hline
Adaptive   & \ding{55} & \ding{55}  & \textbf{\checkmark} & \textbf{\checkmark}      \\ \midrule
RMSE       & 4.87   & 4.88   & 3.84   & 3.67   \\ \bottomrule 
\end{tabular}
\end{center}
\end{table}
\setlength{\tabcolsep}{1.4pt}

\begin{figure}[!t]
  \centering
   \includegraphics[width=0.52\linewidth]{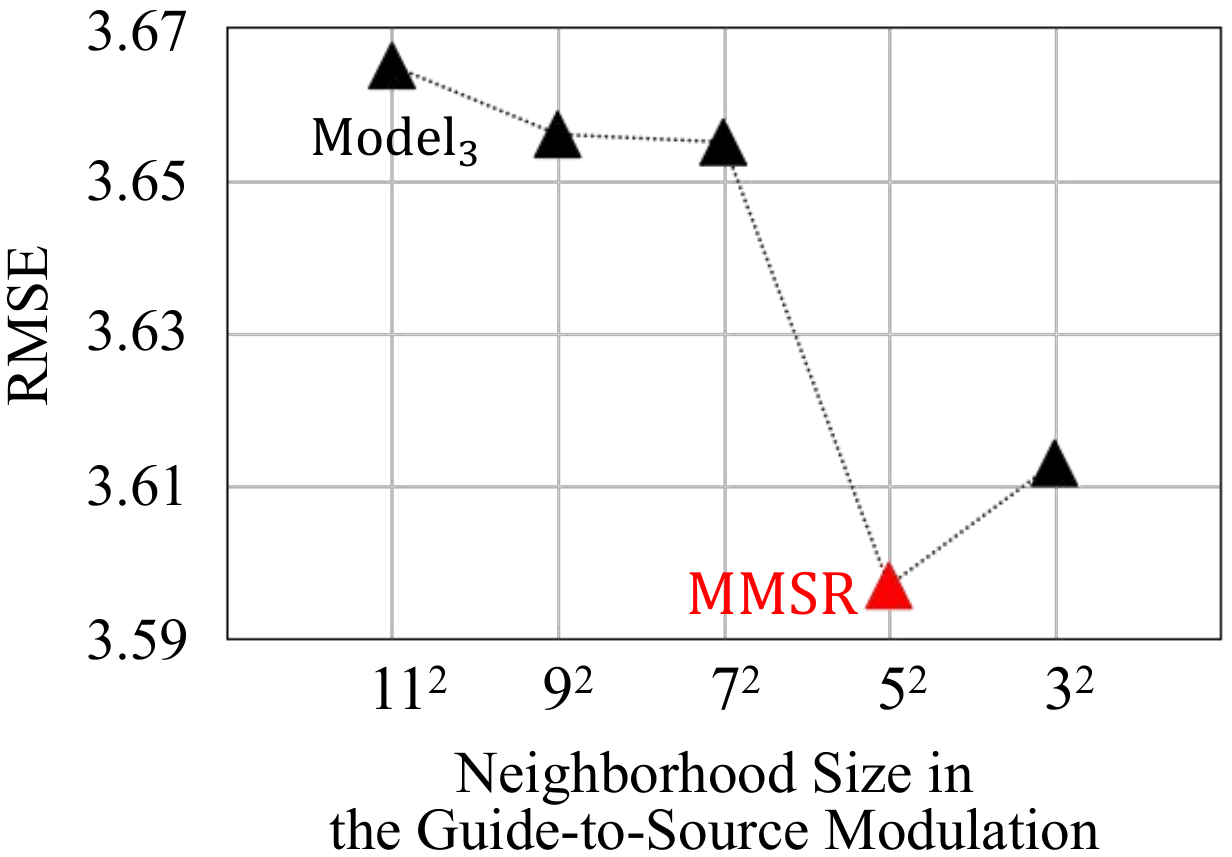}
   \caption{Effect of asymmetric neighborhood sizes. 
   The neighborhood in the source-to-guide modulation is fixed as $11\times 11$}
   \label{fig10_reduceg2s}  
\end{figure}

\noindent\textbf{Effect of Asymmetric Neighborhood Sizes.}    
As introduced in~\cref{sec_modulation}, our mutual modulation is driven by filtering confined neighborhoods in the source and guide features. 
In general image restoration, properly increasing filtering size benefits model performance~\cite{Liu_2018_NLRN}. 
In our experiments, to Model$_3$ in~\Cref{table2_modulation},~if the neighborhood sizes in the source-to-guide and guide-to-source modulations are both set as $3\times 3$ or $7\times 7$, the obtained RMSE values are correspondingly $4.23$ and $3.91$.  
When both are increased to $11\times 11$, as in~\Cref{table2_modulation}, the RMSE is $3.67$.
Considering GPU memory limitations, we did not increase the sizes further. 
Since our modulation strategy is bidirectional, we further investigate the effect of asymmetric neighborhood sizes. 
Based on Model$_3$, we fix the neighborhood size in the source-to-guide modulation as $11\times 11$, while reducing that in the guide-to-source modulation, as shown in~\cref{fig10_reduceg2s}.  
The performance peaks at $5\times 5$, which shows that there is an optimal setting on specific types of image data.  
On the Middlebury data, 
when the neighborhood sizes in the source-to-guide and guide-to-source modulations are respectively $11\times 11$ and $5\times 5$, 
our model can modulate the source and the guide optimally.  
Therefore, we adopt this setting to our model in~\Cref{subsec:comparison}.
When we fixed the neighborhood in the guide-to-source modulation and reduced that in the source-to-guide modulation, the results were overly decided by the source. 
Visual results and more analyses of these two cases are in the supplementary material.  

%

\subsection{Validation on Real-World DEM and Thermal}  
\label{subsec:generalizability}

\setlength{\tabcolsep}{4pt}
\begin{table} [t]
\begin{center}
\caption{DEM SR evaluation by observing average RMSE. The best and the second best results are in \textcolor{red}{red} and \textcolor{blue}{blue}, respectively}
\label{table4_dem}
\begin{tabular}{c|ccc|ccc}
\toprule
          & \multicolumn{3}{c|}{Supervised} & \multicolumn{3}{c}{Self-Supervised}          \\ \hline 
Scale     & DKN~\cite{Kim_2021_IJCV} & FDKN~\cite{Kim_2021_IJCV} & FDSR~\cite{He_2021_CVPR} & P2P~\cite{Lutio_2019_ICCV} & CMSR~\cite{Shacht_2021_CVPR} & MMSR~(Ours) \\ \hline
$\times4$ & 0.80 & 0.80                        & 0.81  & 1.57 & {\color[HTML]{3531FF} 0.78}& {\color[HTML]{FE0000} 0.73} \\ \hline
$\times8$ & 1.39 & {\color[HTML]{3531FF} 1.25} & 1.55 & 1.70 & - & {\color[HTML]{FE0000} 1.02} \\ \bottomrule 
\end{tabular}
\end{center}
\end{table}
\setlength{\tabcolsep}{1.4pt}

\begin{figure*}[t]
  \centering
  \includegraphics[width=1\linewidth]{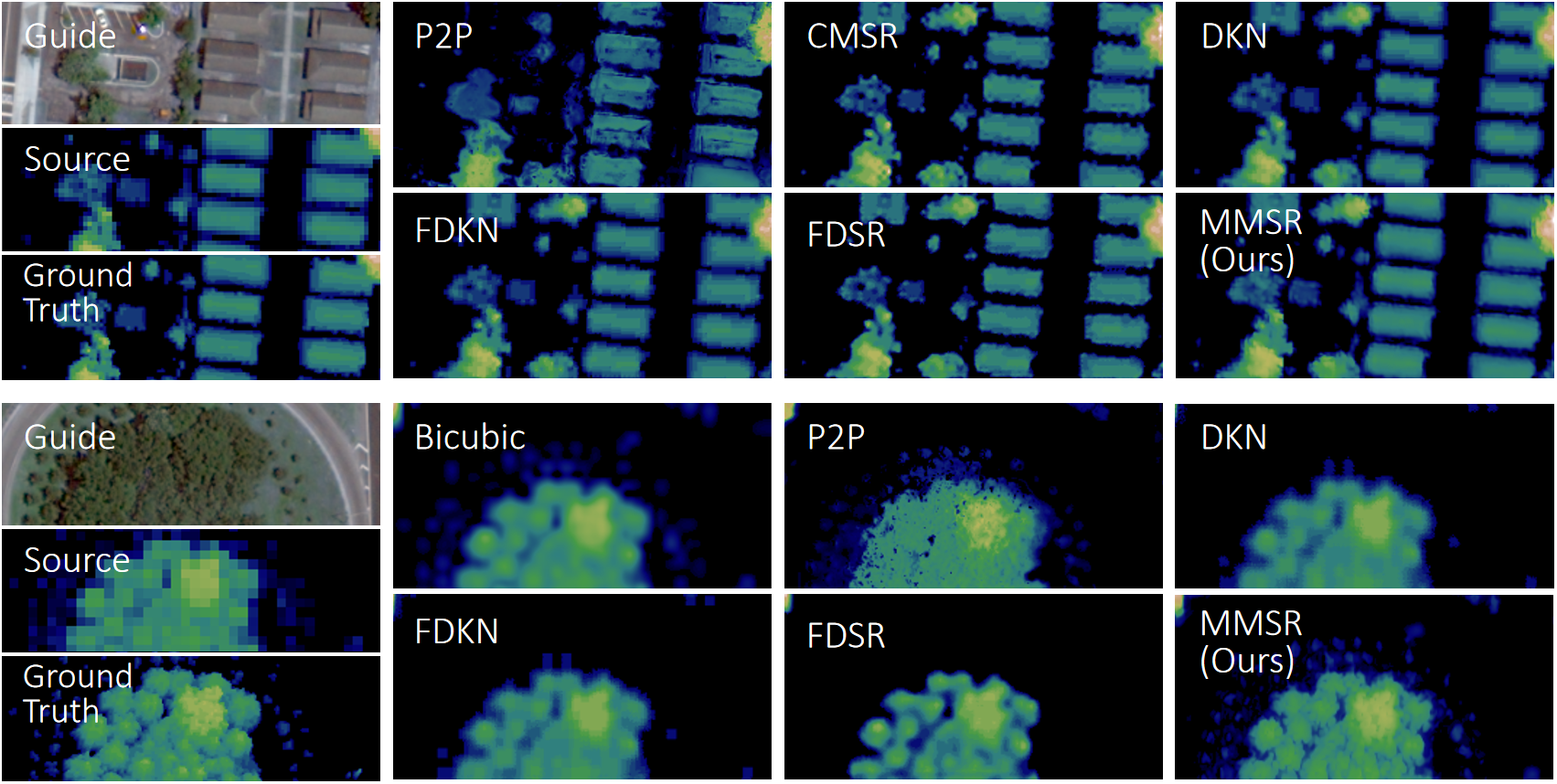} 
  \caption{DEM SR. The upper and lower rows show $\times 4$ and $\times 8$ SR results, respectively}
  \label{fig11_dem}
\end{figure*} 

Given the importance of SR techniques in Earth observation,
we apply our MMSR to real-world remote sensing data that covers DEM and thermal modalities.  
The neighborhood sizes in the source-to-guide and guide-to-source modulations are set as $5 \times 5$ and $3 \times 3$, respectively. 

We compare MMSR with the five state-of-the-art methods~\cite{Shacht_2021_CVPR,He_2021_CVPR,Kim_2021_IJCV,Lutio_2019_ICCV} in terms of $\times 4$ and $\times 8$ DEM SR.
As reported in~\Cref{table4_dem}, our MMSR yields the best quantitative performance under both scale factors, and outperforms the three supervised methods by a large margin.
It can be seen qualitatively from~\cref{fig11_dem}~that our MMSR shows superiority by preserving finer spatial details for the buildings and plants and the modality characteristics of DEM. 
 
\begin{figure}[t]
  \centering
   \includegraphics[width=1\linewidth]{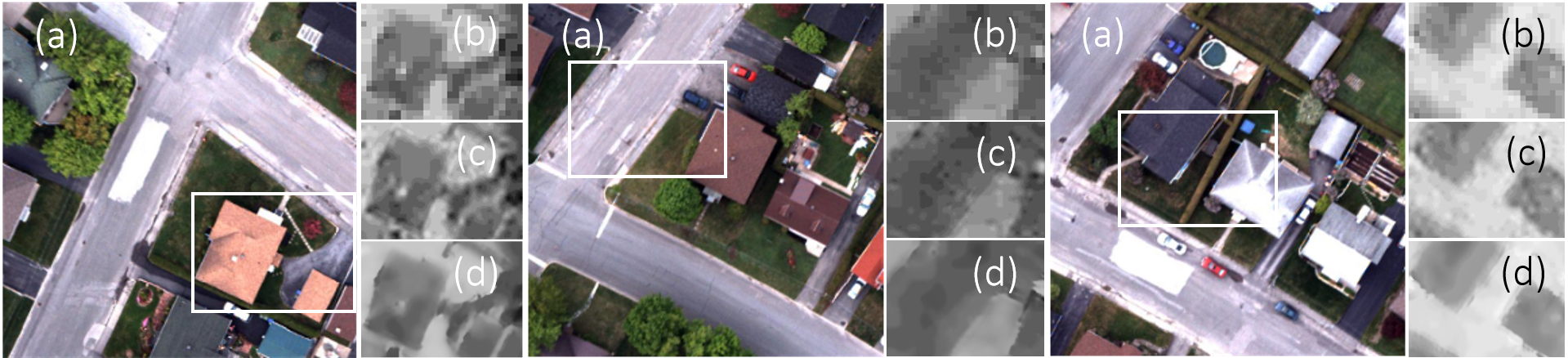} 
   \caption{$\times 5$ thermal SR. (a) Guide. (b) Source. 
   Results from (c)~CMSR and (d)~Ours  
   }  
   \label{fig12_thermal}
\end{figure}

We further compare MMSR with CMSR~\cite{Shacht_2021_CVPR} on the visible-thermal data from~\cite{DFC_2014_thermal}. 
We do not provide numerical evaluation since only LR thermal data is available. 
As presented in \cref{fig12_thermal}, our MMSR shows robust performance and superior generalizability. 
More visual results are in the supplementary material.  

\section{Conclusions}

We study cross-modal SR and present a robust self-supervised MMSR model.
Within MMSR, 
we introduce a mutual modulation strategy to overcome the LR problem of the source and the discrepancy problem of the guide, 
and adopt a cycle consistency constraint to conduct training in a fully self-supervised manner, without accessing ground truth or external training data. 
We demonstrate the superior generalizability of our MMSR on depth, DEM, and thermal modalities,  
and its applicability to noisy data and real-world remote sensing data.
Extensive experiments demonstrate the state-of-the-art performance of our MMSR. 

We believe our concept of modulating different modalities to achieve self-supervised cross-modal SR can inspire further progress in this field, and believe our MMSR can contribute to Earth observation applications where images in various modalities are available but HR ones are rare and expensive. \\


\noindent\textbf{Acknowledgements.}  
XD was supported by the RIKEN Junior Research Associate (JRA) Program.  
NY was was supported by JST, FOREST Grant Number JPMJFR206S, Japan.



\clearpage
%
%
\bibliographystyle{splncs04}



\renewcommand \thetable{\Roman{table}}             
\renewcommand \thefigure{\Roman{figure}}
\renewcommand \thesection{\Roman{section}}
\renewcommand \thesubsection{\Roman{section}.\Roman{subsection}}


\pagestyle{headings}
\mainmatter
\def\ECCVSubNumber{4336}  

\title{Supplementary Material: \\ Learning Mutual Modulation for \\ Self-Supervised Cross-Modal Super-Resolution} 


\titlerunning{Learning Mutual Modulation for Self-Supervised Cross-Modal SR}
%
\author{Xiaoyu Dong\inst{1,2} \and   
Naoto Yokoya\inst{1,2}$^{(\textrm{\Letter})}$ \and
Longguang Wang\inst{3} \and
Tatsumi Uezato\inst{4}}
\authorrunning{X. Dong et al.}
%
\institute{The University of Tokyo, Tokyo, Japan \and 
RIKEN AIP, Tokyo, Japan \and
National University of Defense Technology, Changsha, China \and
Hitachi, Ltd, Tokyo, Japan \\
\email{dong@ms.k.u-tokyo.ac.jp, yokoya@k.u-tokyo.ac.jp} \\
\url{https://github.com/palmdong/MMSR}}
\maketitle


\Cref{sec:asymmetric_effect} analyzes mutual modulation with asymmetric neighborhood sizes.
\Cref{sec:fusion_approaches} studies different feature fusion approaches.    
\Cref{sec:self_methods} compares the time cost of different self-supervised cross-modal super-resolution (SR) methods, and further compares their performance under noisy guidance.
\Cref{sec:discussions} provides more discussions.    
\Cref{sec:more_results} provides more qualitative results.  


\section{Modulation with Asymmetric Neighborhood Sizes}  
\label{sec:asymmetric_effect}

In Section~4.3 Ablation Study, we have discussed the effect of the asymmetric neighborhood sizes in our mutual modulation. 

\begin{figure*}[!h]
  \centering
  \includegraphics[width=0.76\linewidth]{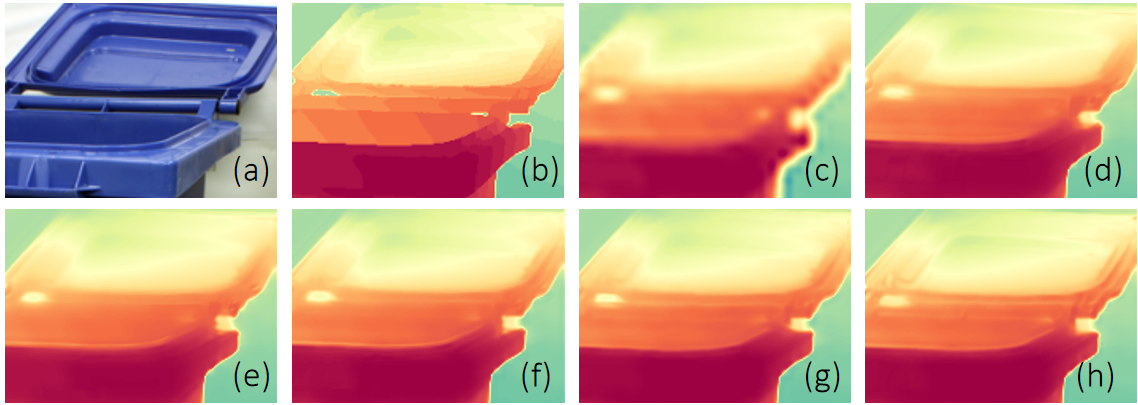} 
  \caption{
  (a) Guide. (b) Ground truth. (c) Bicubic source. Results from models of which the neighborhood sizes for the guide-to-source modulation are (d) $11\times 11$, (e)~$9\times 9$, (f) $7\times 7$, (g) $5\times 5$, and (h) $3\times 3$, respectively}     
  \label{supp_reduceg2s}
\end{figure*}

\begin{figure*}[!h]
  \centering
   \includegraphics[width=0.53\linewidth]{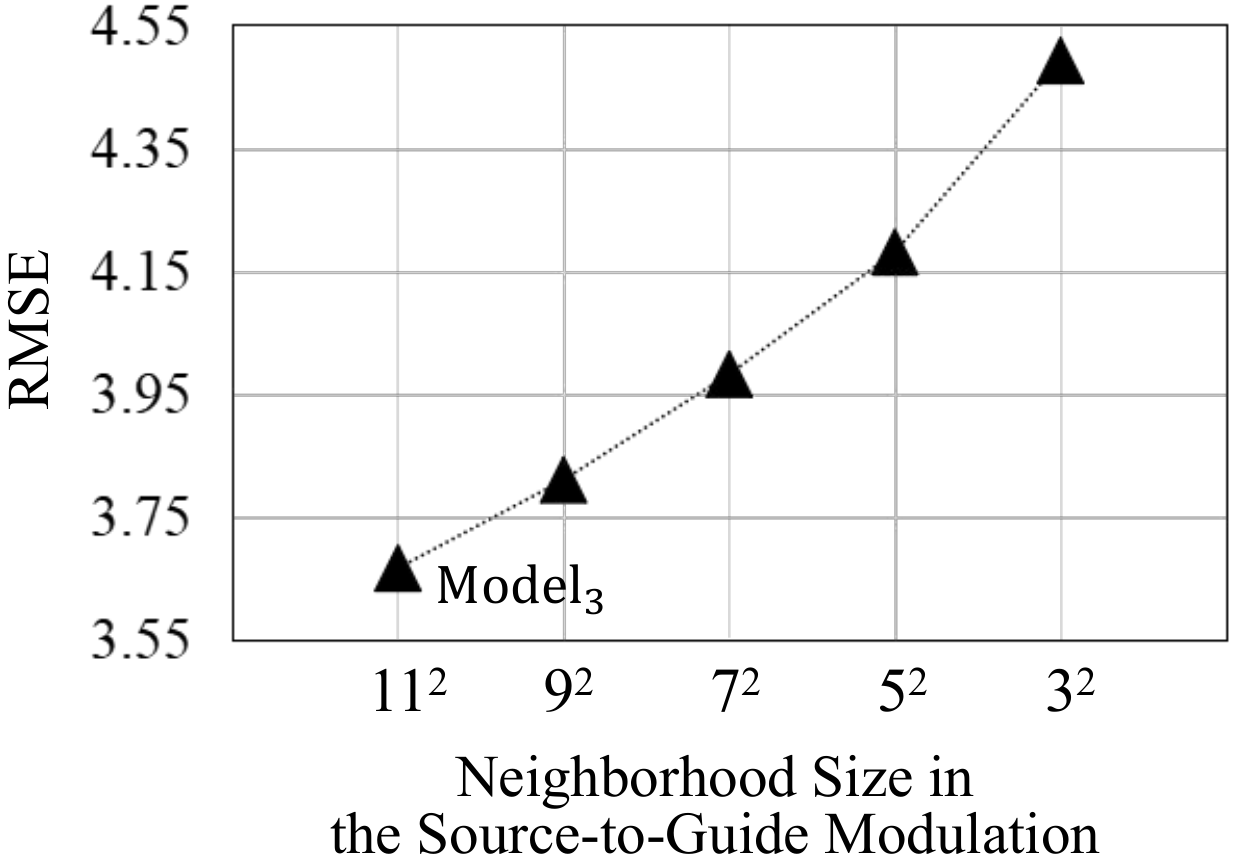}  
   \caption{Effect of asymmetric neighborhood sizes. The neighborhood in the guide-to-source modulation is fixed as $11\times 11$}  
   \label{supp_reduces2g_curve}
\end{figure*}

\begin{figure*}[t]
  \centering
  \includegraphics[width=0.76\linewidth]{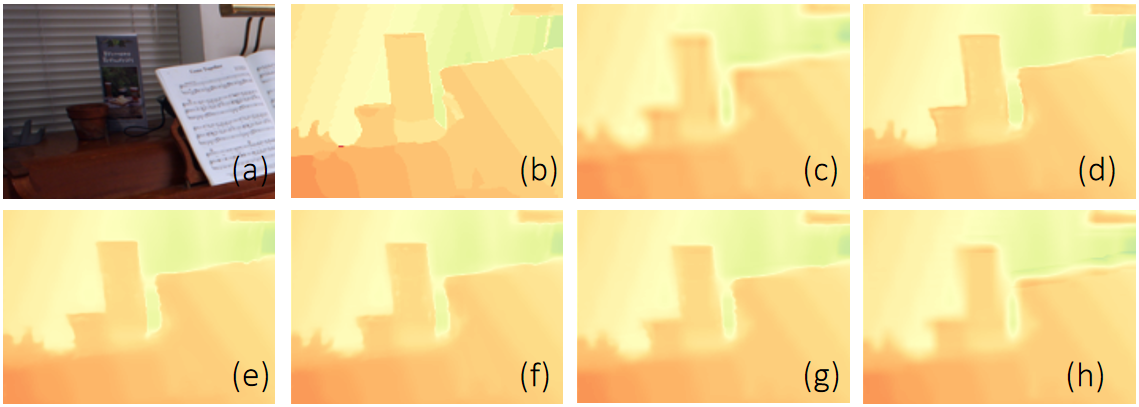}  
  \caption{(a) Guide. (b) Ground truth. (c) Bicubic source. Results from models of which the neighborhood sizes for the source-to-guide modulation are (d) $11\times 11$, (e) $9\times 9$, (f) $7\times 7$, (g) $5\times 5$, and (h) $3\times 3$, respectively}       
  \label{supp_reduces2g}
\end{figure*}

\cref{supp_reduceg2s} provides visual examples of the case in which we fixed the neighborhood size in the source-to-guide modulation as $11\times 11$ and reduced that in the guide-to-source modulation. 
When the neighborhood size is reduced to $9\times 9$,  
the result (\cref{supp_reduceg2s}(e)) is lack of details, because the spatial suppression to the guide is strong.  
When it is further reduced to $3\times 3$, 
the result (\cref{supp_reduceg2s}(h)) has extraneous structures, because the suppression to the spatial discrepancy in the guide is weak.  
When it is set as $5\times 5$, 
the result (\cref{supp_reduceg2s}(g)) is close to the ground truth (\cref{supp_reduceg2s}(b)) with respect to both spatial resolution and modality characteristics.

\cref{supp_reduces2g_curve} reports the quantitative results of the case in which we fixed the neighborhood size in the guide-to-source modulation as $11\times 11$ and reduced that in the source-to-guide modulation.
\cref{supp_reduces2g}~provides visual examples. 
As the neighborhood size reduces, the results become blurry.
This is because the strength to increase the resolution of the source is reduced.

In summary, our mutual modulation allows to handle different types of multi-modal data flexibly.
With setting a large neighborhood size for the source-to-guide modulation and a properly small neighborhood size for the guide-to-source modulation, models can optimally increase the resolution of the source and capture and suppress the spatial discrepancy in the guide. 


\section{Different Feature Fusion Approaches} 
\label{sec:fusion_approaches}

\setlength{\tabcolsep}{4pt}
\begin{table}[!h]   
\begin{center}
\caption{$\times 4$ depth SR on the Middlebury 2003 dataset}  
\label{table_fusion}
\begin{tabular}{c|c|c|c|c}
\toprule
     & ~~~~~~Sum.~~~~~~ & ~Att. $+$ Sum.~ & Att. $+$ Conv$_{1\times1}$ & ~~~~Conv$_{1\times1}$~~~~ \\ \hline
RMSE & 1.88     & 1.92 & 1.84  & \textbf{1.78} \\
Training Time    & ~140s     & ~150s   & ~147s    & \textbf{~137s}  \\ \bottomrule 
\end{tabular}
\end{center}
\end{table}
\setlength{\tabcolsep}{1.4pt}

In MMSR, the modulated features $\textbf{F}_{s2g}$ and $\textbf{F}_{g2s}$ are fused by a $1\times1$ convolution.
We additionally studied other fusion approaches, including naive summation and attentional fusion (spatial attention and channel attention were performed before summation or $1\times 1$ convolution), as reported in \Cref{table_fusion}.
The model with only a $1\times 1$ convolution as fusion approach achieves the best performance and the shortest training time.   
Therefore, a $1\times 1$ convolution is adopt to fuse the modulated features in our MMSR.  


\section{Comparisons with Other Self-Supervised Methods}    
\label{sec:self_methods}

\noindent\textbf{Time Cost.}
As introduced in Section~1, self-supervised cross-modal SR methods, including CMSR~\cite{Shacht_2021_CVPR}, P2P~\cite{Lutio_2019_ICCV}, and our MMSR, perform online learning on each combination of low-resolution (LR) source and high-resolution (HR) guide. 
\Cref{table_time} compares their training time cost. 
Note that, the time cost of our MMSR is influenced by the modulation neighborhood sizes (i.e., modulation with larger neighborhood sizes results in higher time cost).  
For depth SR,  
the neighborhood sizes for the source-to-guide modulation and the guide-to-source modulation in our MMSR were set as $11\times 11$ and $5\times 5$, respectively.  
The training time of MMSR/P2P/CMSR on each depth-visible input (of size $320\times 320$) is 137s/131s/90s.
Our MMSR runs slightly slower yet shows obvious performance superiority, as shown in~\Cref{table_time}.  
For digital elevation model~(DEM) SR,  
the neighborhood sizes for the source-to-guide modulation and the guide-to-source modulation in our MMSR were set as $5\times 5$ and $3\times 3$, respectively. 
The training time of MMSR/P2P/CMSR on each DEM-visible input (of size $320\times 320$) is 49s/131s/90s.
Our MMSR requires much less time and still obtains obvious performance superiority. 

\setlength{\tabcolsep}{4pt}
\begin{table}[!h]
\begin{center}
\caption{$t$ shows training time on an NVIDIA RTX 3090 GPU. RMSE$_{2003}$, RMSE$_{2005}$, and RMSE$_{2014}$ denote the average RMSE on the Middlebury 2003~\cite{Scharstein_2003_CVPR}, 2005~\cite{Scharstein_2007_CVPR}, and 2014~\cite{Scharstein_2014_GCPR} datasets, respectively. Numbers in brackets show the performance improvement achieved by our MMSR}          
\label{table_time}
\begin{tabular}{c|c|ccc|c|c}
\toprule
    & \multicolumn{4}{c|}{$\times 4$ Depth SR} & \multicolumn{2}{c}{$\times 4$ DEM SR}          \\ \hline
    & $t$ & RMSE$_{2003}$    & RMSE$_{2005}$   & RMSE$_{2014}$    & $t$ & RMSE  \\ \midrule
P2P~\cite{Lutio_2019_ICCV} & 131s & 2.94 {\scriptsize ($\uparrow 39.5\%$)} & 3.78 {\scriptsize($\uparrow 34.7\%$)}  & 3.90 {\scriptsize($\uparrow 41.0\%$)}   & 131s & 1.57 {\scriptsize($\uparrow 53.5\%$)} \\ \hline
CMSR~\cite{Shacht_2021_CVPR} & 90s & 2.52 {\scriptsize($\uparrow 29.4\%$)} & 3.51 {\scriptsize($\uparrow 29.6\%$)}  & 2.87 {\scriptsize($\uparrow 19.9\%$)}  & 90s & 0.78 {\scriptsize($\uparrow 6.4\%$)} \\ \hline
Ours & 137s & 1.78 {\scriptsize(-)} & 2.47 {\scriptsize(-)} & 2.30 {\scriptsize(-)}                            & 49s & 0.73 {\scriptsize(-)} \\ \bottomrule 
\end{tabular}
\end{center}
\end{table}
\setlength{\tabcolsep}{1.4pt}

\begin{figure*}[h]
  \centering
  \includegraphics[width=1\linewidth]{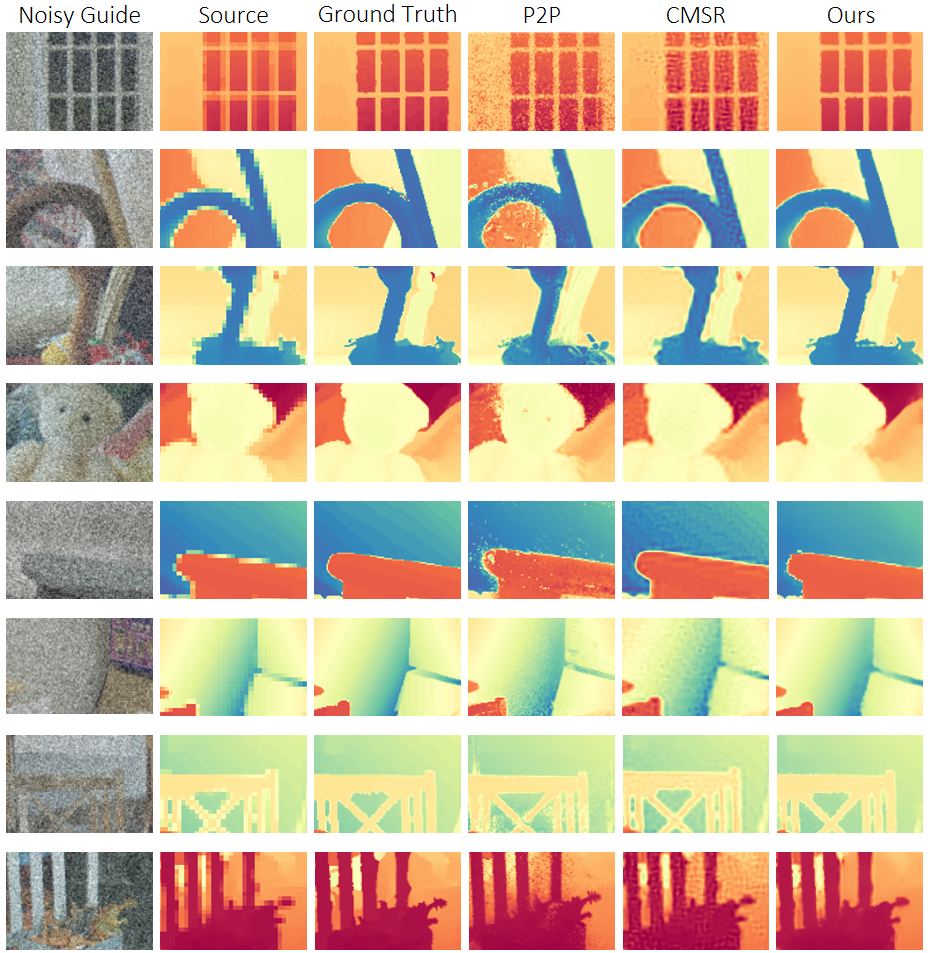} 
  \caption{
  $\times 4$ depth SR under noisy guidance. The first four and the second four rows show results on the Middlebury 2005 and 2014 datasets, respectively. $'$Noisy Guide$'$ is generated by adding Gaussian noise with noise level 50
  }     
  \label{supp_noisy}
\end{figure*}

\noindent\textbf{Performance under Noisy Guidance.}   
\cref{supp_noisy} further compares our MMSR with CMSR~\cite{Shacht_2021_CVPR} and P2P~\cite{Lutio_2019_ICCV} under noisy guidance.
As we can see, under even heavy noise, our MMSR still outperforms CMSR and P2P by a large margin and can produce results that are closer to ground truth. This demonstrates the robustness of our MMSR and the effectiveness of our mutual modulation with cross-domain adaptive filtering.   

\section{More Discussions}    
\label{sec:discussions}

\noindent\textbf{What Is Important in Cross-Modal SR?}   
Given an LR source and an HR guide from different modalities, cross-modal SR aims at achieving an image product that has spatial resolution comparable with the guide and modality characteristics faithful to the source. 
We argue both the structural cues from the HR guide and the modality constraint from the LR source are important in the task.
Thus we develop a mutual modulation strategy and adopt cycle consistency constraint to fully exploit the guide and also the source, enabling a robust self-supervised MMSR model. 

\noindent\textbf{Why Can MMSR Outperform Supervised Methods?}
Supervised cross-modal SR methods have shown promising performance.    
However, they have two problems:
\textbf{(1)}~They suffer limited performance in real-world scenes because large-scale paired training data is hard to acquire. 
\textbf{(2)}~They cannot easily generalize well to test data that is not in the same domain as the training data. 
The reasons of our superior performance are twofold: 
\textbf{(1)}~Our mutual modulation strategy and cycle-consistent self-supervised learning effectively facilitate our MMSR to achieve state-of-the-art performance.
\textbf{(2)}~The employed online learning scheme allows our MMSR a strong generalization capability to any given input.
With robust performance and strong generalizability, MMSR can outperform even supervised methods.

\noindent\textbf{Contributions beyond Superior Performance.}
Our MMSR outperforms previous supervised and self-supervised methods on various tasks.
Moreover, our work also has the following three major contributions: 
\textbf{(1)}~The state-of-the-art performance of our MMSR bridges the gap of robust self-supervised cross-modal SR.   
\textbf{(2)}~For the first time, our mutual modulation effectively overcomes the spatial discrepancy and resolution gap of multi-modal images, and show correlation-based filtering provides an effective inductive bias for deep cross-modal SR. 
This benefits further progress in research fields. 
\textbf{(3)}~Our MMSR shows superior generalization capability to diverse modalities, robustness to noise, and applicability to real-world scenarios. 
This is beneficial to real-world applications.  

\noindent\textbf{Limitation.} 
Like other methods, MMSR produces ghosting artifacts on some samples. 
In Fig. 5, ghosting artifacts can be observed around the antlers in the results of FDSR~\cite{He_2021_CVPR}, FDKN~\cite{Kim_2021_IJCV}, DKN~\cite{Kim_2021_IJCV}, and MMSR.
This is caused by the bicubic/bilinear upsampled source input.
Since P2P~\cite{Lutio_2019_ICCV} inputs only the guide image, it does not suffer from ghosting artifacts but produces discrepancy artifacts.  
Likewise, in~Fig. 9, in feature $\textbf{F}_{g2s}$, the ghosting along the antler is because the guide-to-source modulation induces $\textbf{F}_{g}$ to mimic $\textbf{F}_{s}$ which has bilinear ghosting. 
However, compared with previous state-of-the-art methods~\cite{He_2021_CVPR,Shacht_2021_CVPR,Kim_2021_IJCV,Lutio_2019_ICCV}, our MMSR achieves final predictions that are closer to ground truth.  
Exploring the upperbound performance of self-supervised cross-modal SR models would be an interesting and challenging research problem.


\section{More Qualitative Results}   
\label{sec:more_results}

\begin{figure*}[h]
  \centering
  \includegraphics[width=1\linewidth]{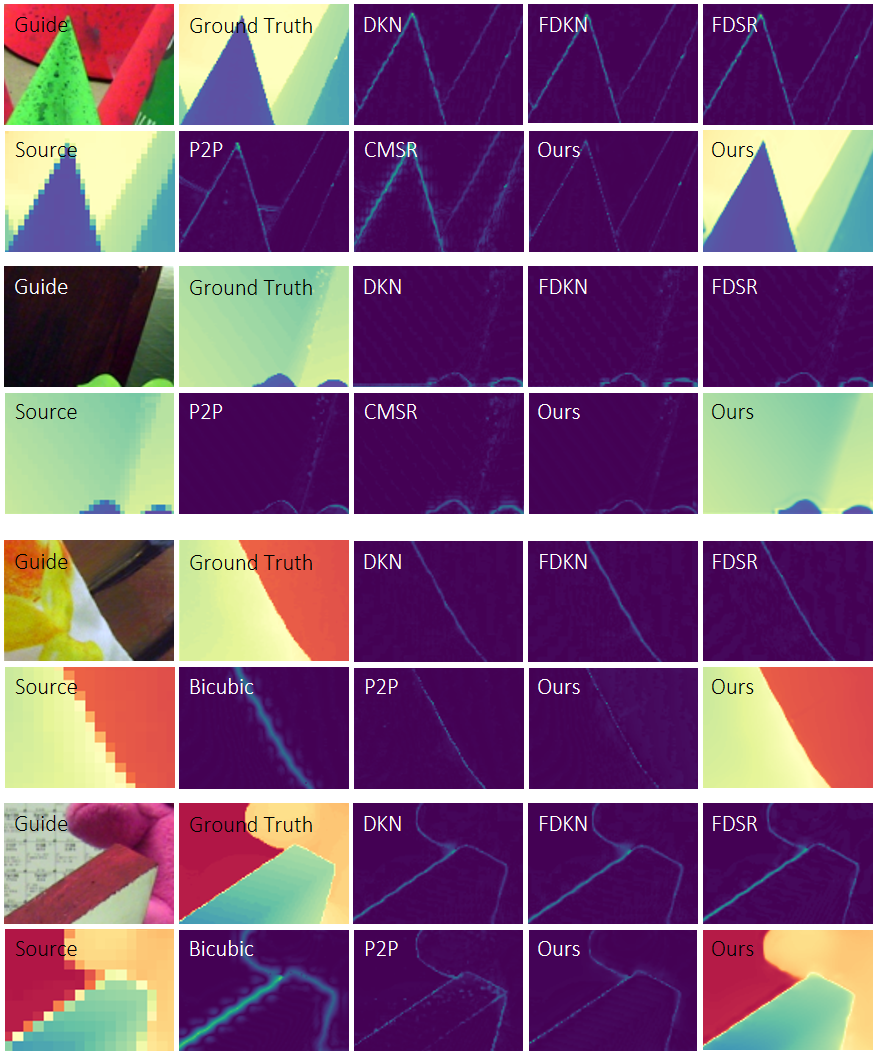} 
  \caption{
  Depth SR on the Middlebury 2003 dataset.
  The first and second rows show $\times 4$ SR results. 
  The third and fourth rows show $\times 8$ SR results.
  }     
  \label{supp_depth_2003}
\end{figure*}

\begin{figure*}[h]
  \centering
  \includegraphics[width=1\linewidth]{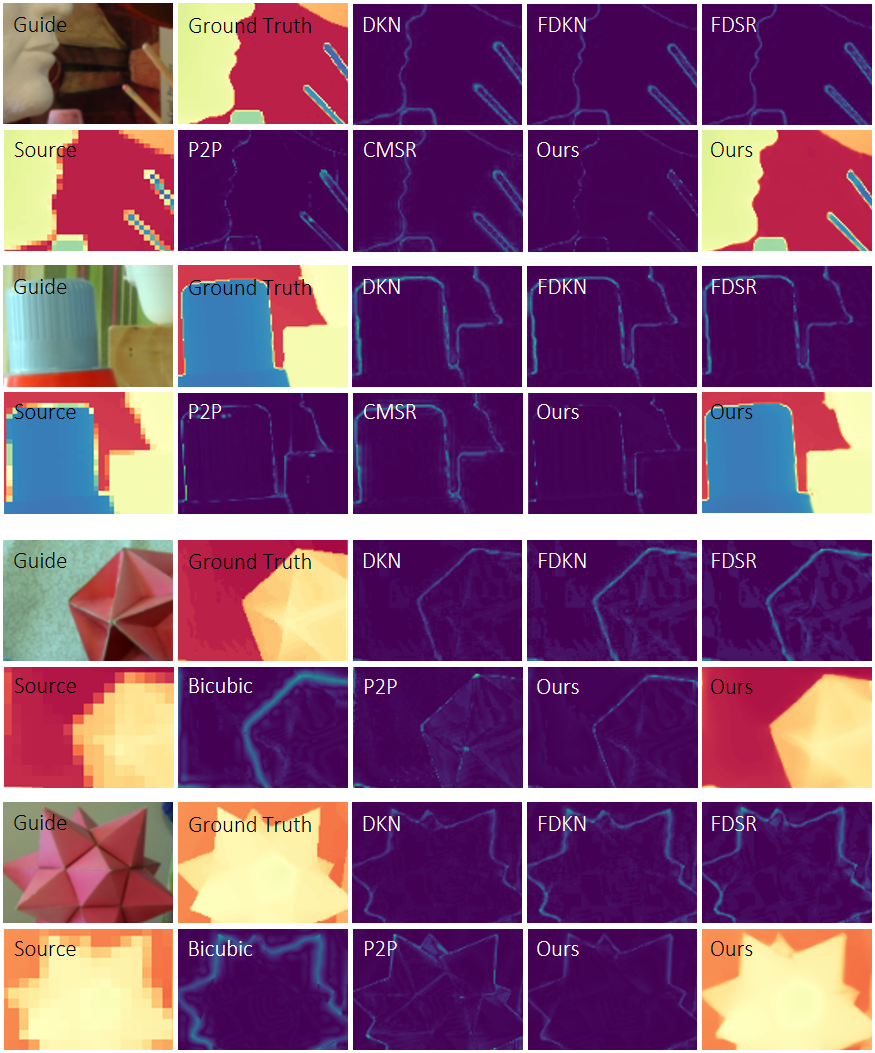}  
  \caption{
  Depth SR on the Middlebury 2005 dataset.
  The first and second rows show $\times 4$ SR results. 
  The third and fourth rows show $\times 8$ SR results
  }     
  \label{supp_depth_2005}
\end{figure*}

\begin{figure*}[h]
  \centering
  \includegraphics[width=1\linewidth]{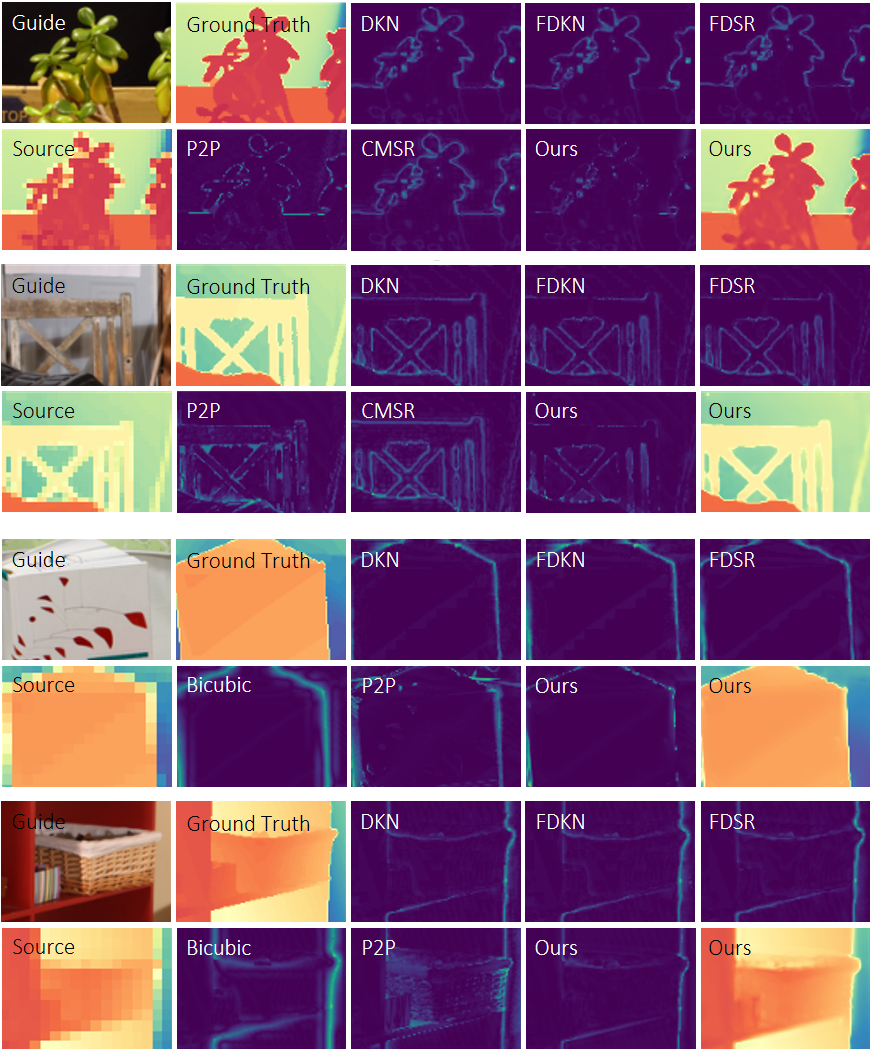} 
  \caption{
  Depth SR on the Middlebury 2014 dataset.
  The first and second rows show $\times 4$ SR results. 
  The third and fourth rows show $\times 8$ SR results.
  }     
  \label{supp_depth_2014}
\end{figure*}

\begin{figure*}[t]
  \centering
  \includegraphics[width=0.98\linewidth]{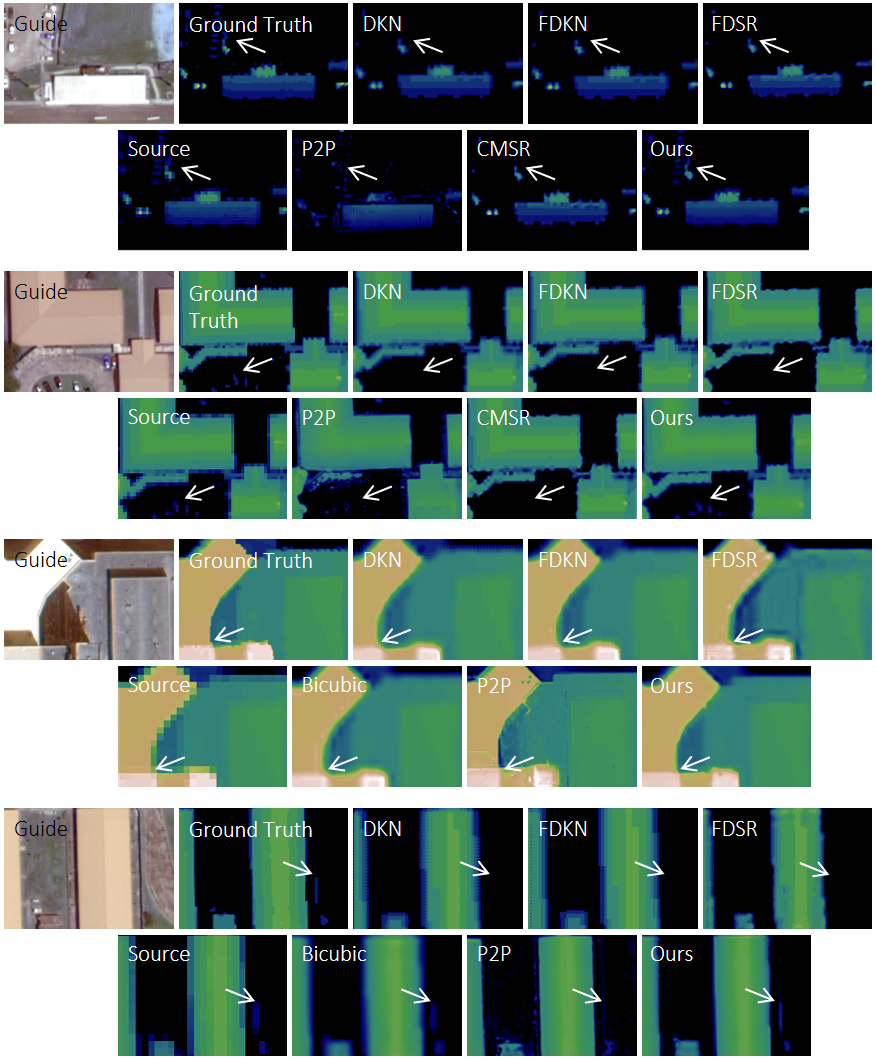}
  \caption{DEM SR. The first and second rows show $\times 4$ SR results. The third and fourth rows show $\times 8$ SR results 
  }     
  \label{supp_dem}
\end{figure*}

We provide more visual comparisons between our MMSR and the five cross-modal SR methods~\cite{Shacht_2021_CVPR,He_2021_CVPR,Kim_2021_IJCV,Lutio_2019_ICCV}.  
\cref{supp_depth_2003}, \cref{supp_depth_2005}, and \cref{supp_depth_2014} show SR results on the depth-visible data from the Middlebury 2003~\cite{Scharstein_2003_CVPR}, 2005~\cite{Scharstein_2007_CVPR}, and 2014~\cite{Scharstein_2014_GCPR} benchmarks, respectively.  
\cref{supp_dem} shows SR results on the real-world DEM-visible data from~\cite{DFC_2019_DEM}.  
For depth SR, error maps are provided for better visual comparison. 
As we can see, our MMSR produces lower errors and finer edge details. 
Overall, as a self-supervised method, our MMSR achieves state-of-the-art performance on various tasks, and outperforms fully supervised methods (FDSR~\cite{He_2021_CVPR}, DKN~\cite{Kim_2021_IJCV}, and FDKN~\cite{Kim_2021_IJCV}) and previous self-supervised methods (CMSR~\cite{Shacht_2021_CVPR} and P2P~\cite{Lutio_2019_ICCV}) consistently.

%
%
\end{document}